\documentclass[journal]{IEEEtran}
\pdfoutput=1
\ifCLASSINFOpdf
\else
\fi

\usepackage{amsfonts}       

\usepackage{amsmath}
\usepackage{amssymb}
\usepackage{graphicx}
\usepackage{float}
\usepackage{graphics}
\usepackage{multirow}
\usepackage{bm}
\usepackage{mathrsfs}
\usepackage{algorithm}
\usepackage{algorithmic}
\usepackage{soul,color}
\usepackage{mathtools}
\usepackage{xcolor}
\usepackage{colortbl}
\usepackage[thinc]{esdiff}

\DeclareMathOperator{\E}{\mathbb{E}}
\DeclareMathOperator{\Var}{\mathrm{Var}}

\def\ddefloop#1{\ifx\ddefloop#1\else\ddef{#1}\expandafter\ddefloop\fi}

\newcommand\abs[1]{|#1|}

\def\ddef#1{\expandafter\def\csname v#1\endcsname{\ensuremath{\boldsymbol{#1}}}}
\ddefloop ABCDEFGHIJKLMNOPQRSTUVWXYZabcdefghijklmnopqrstuvwxyz\ddefloop

\def\ddef#1{\expandafter\def\csname v#1\endcsname{\ensuremath{\boldsymbol{\csname #1\endcsname}}}}
\ddefloop {alpha}{beta}{gamma}{delta}{epsilon}{varepsilon}{zeta}{eta}{theta}{vartheta}{iota}{kappa}{lambda}{mu}{nu}{xi}{pi}{varpi}{rho}{varrho}{sigma}{varsigma}{tau}{upsilon}{phi}{varphi}{chi}{psi}{omega}{Gamma}{Delta}{Theta}{Lambda}{Xi}{Pi}{Sigma}{varSigma}{Upsilon}{Phi}{Psi}{Omega}{ell}\ddefloop

\def\ddef#1{\expandafter\def\csname bb#1\endcsname{\ensuremath{\mathbb{#1}}}}
\ddefloop ABCDEFGHIJKLMNOPQRSTUVWXYZ\ddefloop

\begin{document}
%
\title{A Batched Scalable Multi-Objective Bayesian Optimization Algorithm}
%
%
%

\author{Xi~Lin,
		Hui-Ling Zhen,
        Zhenhua Li,
        Qingfu~Zhang,~\IEEEmembership{Fellow,~IEEE,}
        Sam~Kwong,~\IEEEmembership{Fellow,~IEEE}
\thanks{This work is supported by Hong Kong RGC General Research Fund(GRF) 9042038 (CityU 11205314) and the National Natural Science Foundation of China under Grants 61473241.}
\thanks{X.~Lin, H. L.~Zhen, Z.~Li, Q.~Zhang, and S.~Kwong are with the Department of Computer Science, City University of
Hong Kong, Hong Kong. (email: \{xi.lin,zhenhua.li\}@my.cityu.edu.hk, \{qingfu.zhang, cssamk\}@cityu.edu.hk)}
\thanks{}
}

\maketitle

\begin{abstract}
The surrogate-assisted optimization algorithm is a promising approach for solving expensive multi-objective optimization problems. However, most existing surrogate-assisted multi-objective optimization algorithms have three main drawbacks: 1) cannot scale well for solving problems with high dimensional decision space, 2) cannot incorporate available gradient information, and 3) do not support batch optimization. These drawbacks prevent their use for solving many real-world large scale optimization problems. This paper proposes a batched scalable multi-objective Bayesian optimization algorithm to tackle these issues.  The proposed algorithm uses the Bayesian neural network as the scalable surrogate model. Powered with Monte Carlo dropout and Sobolov training, the model can be easily trained and can incorporate available gradient information. We also propose a novel batch hypervolume upper confidence bound acquisition function to support batch optimization. Experimental results on various benchmark problems and a real-world application demonstrate the efficiency of the proposed algorithm.

\end{abstract}

\begin{IEEEkeywords}
surrogate-assisted evolutionary optimization, Bayesian optimization, batch optimization, expensive multi-objective optimization.
\end{IEEEkeywords}

%
\IEEEpeerreviewmaketitle

\section{Introduction}
%
%
%
%

\IEEEPARstart{E}{xpensive} multiobjective optimization problems can be found in many real-world applications. For example, when building a deep neural network, one may want to maximize its model accuracy and minimize its size and respond speed at the same time~\cite{han2015deep}. Surrogate-assisted multiobjective optimization algorithms are promising for solving these problems~\cite{chugh2017survey,deb2018taxonomy}. Much effort has been made on the development of this kind of algorithms~\cite{emmerich2006single,knowles2006parego,ponweiser2008multiobjective,zhang2010expensive,chugh2016surrogate}. However, three major issues remain challenging:

\begin{itemize}

\item \textbf{Scalability:} Existing surrogate-assisted algorithms cannot scale well as the number of decision variables and the number of data points used for surrogate model building increase. These algorithms usually use Gaussian process (Kriging) as the surrogate model~\cite{chugh2017survey,deb2018taxonomy}, of which the cost of model building is cubic to the number of data points used. Thus, they are only suitable for problems with about 10 decision variables and a few hundreds of function evaluations~\cite{chugh2017survey}.

\item \textbf{Use of Gradient Information:} In many real-world applications such as aerodynamic shape optimization~\cite{jameson2003reduction} and the hyperparameter optimization for deep neural networks~\cite{maclaurin2015gradient,fu2016drmad}, the gradients of the optimization objective functions are available with no or low additional cost. Some preliminary works have shown that utilizing gradient information is beneficial for solving single objective expensive optimization problems~\cite{ahmed2016we,wu2017bayesian}. However, the cost of incorporating gradient information into the Gaussian process is very high~\cite{wu2017exploiting}, which prevents the use of full gradient even for small scale problems with a few decision variables~\cite{ahmed2016we,wu2017bayesian}. To our best knowledge, no algorithm has been proposed for utilizing gradient information to solve multiobjective expensive optimization problems.

\item \textbf{Batch Optimization:} Many real-world applications allow parallel evaluations. By simultaneously evaluating many solutions in batch, one can significantly reduce the overall computational clock time which is crucial when the number of required evaluations is large. Although few algorithms can evaluate multiple solutions in batch~\cite{zhang2010expensive,chugh2016surrogate}, most surrogate-assisted multiobjective algorithms are not designed for batch optimization~\cite{emmerich2006single,knowles2006parego,ponweiser2008multiobjective}, where only one solution can be evaluated at each iteration.
\end{itemize}

To address the above three issues, this paper proposes a batched scalable multi-objective Bayesian optimization algorithm (BS-MOBO) for solving expensive multi-objective optimization problems. In the remainder of this paper, we first describe the background and some challenges for the current surrogate-assisted multiobjective algorithms in Section II. In section III, we detail our proposed algorithm BS-MOBO, of which the main advantages are:

\begin{itemize}
\item Instead of using Gaussian process, BS-MOBO builds scalable Bayesian neural networks with Monte Carlo dropout inference as its surrogate model. Powered with Sobolev training technique, BS-MOBO can efficiently incorporate available gradient information into the model building and hence significantly improve the optimization performance.

\item Using our proposed novel batch hypervolume upper confidence bound (B-HUCB) acquisition function, BS-MOBO can simultaneously select a set of promising solutions for parallel evaluation. By considering all already evaluated solutions in the selection step, BS-MOBO can achieve a good exploration-exploitation trade-off for batch selection.

\end{itemize}
In section IV, we compare the proposed algorithm and its variants with state-of-the-art surrogate-assisted multi-objective optimization algorithms. Section V discusses several potential improvements and concludes this paper.

\section{Background and Challenges}

In this section, we introduce the basic surrogate-assisted optimization framework, and discuss some critical challenges for solving scalable multi-objective optimization problems.

\subsection{Expensive Multi-Objective Optimization Problem}
This paper considers the following multi-objective optimization problem (MOP)~\cite{deb2014multi}:

\begin{eqnarray}
    \label{mop}
    \begin{aligned}
        &\min~\boldsymbol{F(x)} = (f_1(\boldsymbol{x}), \ldots, f_m(\boldsymbol{x})) \in \mathbb{R}^m\\
        &s.t.~~~ \vx \in \Omega \subset \mathbb{R}^n
    \end{aligned}
\end{eqnarray}
where $\Omega$ is the decision space, $\mathbb{R}^m$ is the objective space and $\boldsymbol{F(x)}$ is the objective vector. Since the objective functions conflict each other, no single solution can optimize all objectives at the same time. A decision maker is often interested in best trade-off solutions, which is defined by the Pareto optimality~\cite{zitzler1999multiobjective}.

Let $\vx^a,\vx^b \in \mathbb{R}^n$ be two solutions in $\Omega$, $\vx^a$ is said to dominate $\vx^b$ ($\vx^a \prec \vx^b$) if and only if $f_i(\vx^a) \leq f_i(\vx^b), \forall i \in \{1,...,m\}$ and $f_j(\vx^a) < f_j(\vx^b), \exists j \in \{1,...,m\}$. $\vx^{\ast}$ is a Pareto optimal solution and $F(\vx^{\ast})$ is a Pareto optimal objective vector if there is no $\hat \vx \in  \Omega$ such that $\hat \vx \prec \vx^{\ast}$.
The set of all Pareto optimal solutions is called the Pareto set and the set of their corresponding objective vectors is called the Pareto front~\cite{zitzler1999multiobjective}.  The goal of multiobjective optimization algorithms is to find a set of solutions to approximate the Pareto front.

\subsection{Surrogate-Assisted Multi-Objective Optimization}

Surrogate-assisted multi-objective optimization algorithm, such as multi-objective Bayesian optimization (MOBO)\cite{shahriari2016taking,feliot2017bayesian}, is promising for solving expensive multi-objective optimization problems. A framework of MOBO is presented in \textbf{Algorithm~\ref{model_based_opt}}.

      \begin{algorithm}[H]%
      \caption{Multi-objective Bayesian Optimization (MOBO)}
      \label{model_based_opt}
      \begin{algorithmic}[1]
      \STATE Initialize the dataset $D_0 = \{(\vx_i,\vF(\vx_i))| i = 1,\cdots,N_I\}$
      \FOR{$t = 1$ to $T$}
      \STATE build probabilistic surrogate models $\{\hat f_j(\vx|D_{t-1})\}_{j=1}^{m}$%
      \STATE find solution $\vx_t$ by optimizing $\alpha(\vx|D_{t-1})$%
      \item[]\hspace*{0pt}\hfill$\vx_t = \arg\max_{\vx} \alpha(\vx|D_{t-1})$\hfill\hspace*{0pt}%
      \STATE evaluate $\vF(\vx_t)$
      \STATE update $D_{t} = D_{t-1} \cup \{(\vx_t,\vF(\vx_t))\}$%
      \ENDFOR
      \end{algorithmic}
      \end{algorithm}

MOBO first uses an experimental design method to generates a set of $N_I$ solutions and evaluate all of them to obtain the initial dataset $D_0$. At each iteration $t$, MOBO builds a probabilistic surrogate model to approximate each objective function based on the current dataset $D_{t-1}$. With all $m$ surrogate models $\{\hat f_j(\vx|D_{t-1})\}_{j=1}^{m}$, we can define an acquisition function $\alpha(\vx|D_{t-1})$ to measure the gain for evaluating a solution $x$. Expected Hypervolume Improvement (EHI)~\cite{wagner2010expected} is a well-known acquisition function for solving expensive MOP. By optimizing the acquisition function, MOBO locates the most promising solution $\vx_t = \arg\max_{x} \alpha(\vx|D_{t-1})$ and evaluates $\vF(\vx_t)$. At the end of each iteration, the dataset will be augmented by the newly evaluated solution.

The surrogate-assisted optimization algorithm can efficiently solve some expensive MOPs with a limited evaluation budget. However, most existing surrogate-assisted optimization algorithms  have the following drawbacks:
\begin{itemize}
\item These algorithms can only solve small scale problems with a few decision variables since the cost for the surrogate model building is very costly. It is also very difficult to incorporate available gradient information into model building.
\item Most surrogate-assisted algorithms do not support batch optimization, which further weakens its use for solving large-scale optimization problems, especially when the number of required function evaluations is large.
\end{itemize}

We will discuss these challenges in the following subsections and propose an efficient algorithm to tackle them in section III.

\subsection{Scalable Surrogate Model Building}

Gaussian process (Kriging) \cite{rasmussen2006gaussian} is a popular way for building a probabilistic surrogate model. However, the computational cost for training a Gaussian process model is $O(N^3)$ for $N$ training samples. This high computational complexity prevents its use for large scale problems which require a large number of training data~\cite{wu2017bayesian}. Many different choices of surrogate models have been proposed for surrogate-assisted EAs~\cite{chugh2017survey}. However, the performances of these algorithms are usually worse~\cite{chugh2017survey,palar2016comparative,husain2010enhanced}. The lack of a good uncertainty estimation is a main reason for the inferior performance.

Bayesian neural network, which can be trained with a large number of solutions, is an alternative choice to Gaussian process with good uncertainty estimation~\cite{mackay1992practical,neal2012bayesian}. A neural network model with one single hidden layer can be written as:
\begin{align}
\hat y = \hat f(\vx|\vW) =   h(\vx\vW_1 + \vb_1)\vW_2 + \vb_2
\end{align}
where $\vx$ is the input vector, $h(\cdot)$ is a nonlinear activation function, $\vW = \{\vW_1,\vW_2, \vb_1,\vb_2\}$ are the weight matrices and bias terms, and $\hat y$ is the output of the neural network.

\begin{figure}[h]
    \centering
    \includegraphics[width= 0.45\textwidth]{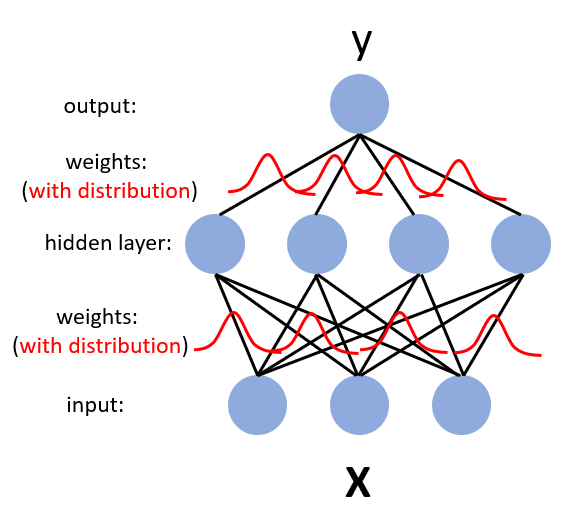}
    \caption{\label{fig:bnn} A Bayesian neural network with one hidden layer.}
\end{figure}

In Bayesian neural network, we place prior distributions $p(\vW)$ on its weights $\vW$ as shown in Fig.~\ref{fig:bnn}. With $N$ training samples $(\vX,\vy) = \{(\vx_i,y_i)| i = 1,\cdots,N\}$, we can obtain the posterior distribution $p(\vW|\vX,\vy)$ on the weight, and hence the output distribution $p(y|\vx,\vW)$~\cite{graves2011practical,blundell2015weight}. The predictive mean and variance of the output can be easily calculated once the output distribution is known. However, the current existing methods for training a Bayesian neural network are usually time consuming and not suitable for many applications~\cite{gal2016dropout}.

Recently, a practical approach which uses dropout as the variational inference method for the Bayesian neural network has been proposed~\cite{gal2016dropout,gal2016uncertainty}. Dropout was originally proposed to avoid over-fitting in training deep neural networks~\cite{hinton2012improving,srivastava2014dropout}. A dropout mask $\vz_i$, which is a binary vector with Bernoulli distribution, is applied for each weight matrix in the neural network:
\begin{align}
&\hat y_{dropout}  =   h(\vx \vZ_1 \overline{\vW}_1 + \vb_1)\vZ_2 \overline{\vW}_2 + \vb_2 \nonumber\\
&\vZ_1 = \text{diag}(\vz_1),  \vZ_2 = \text{diag}(\vz_2)
\end{align}
where diag$(\vz_i)$ maps a vector to a diagonal matrix with $\vz_i$ on its main diagonal. The elements of $\vz_i$ are sampled from a Bernoulli distribution $\vz_{ik} \sim \text{Bernoulli}(p)$ with probability $p$. Here, $\overline{\vW}_i$ is the fixed weight matrix for the neural network to be optimized. The randomness of $\vW_i = \vZ_i \overline{\vW}_i$ only comes from $\vZ_i$. In other words, we randomly zero out some rows in $\overline{\vW}_i$ with probability $1 - p$ by multiplying the dropout mask $\text{diag}(\vz_i)$. For training, the weights of the neural network can be easily optimized by a standard stochastic gradient-based method such as Adam~\cite{kingma2014adam}, of which different dropout masks are sampled at each optimization iteration~\cite{hinton2012improving,srivastava2014dropout}.

For prediction, by randomly sampling the dropout masks $S$ times, we can obtain a set of different dropout outputs $\{\hat y_s(\vx)\}_{s=1}^S$ for a new solution $\vx$. The predictive mean and variance can be estimated empirically:
\begin{align}
   & \E(\hat y) = \frac{1}{S} \sum_{s=1}^{S} \hat y_s(\vx) , \\
   & \Var(\hat y) = \frac{1}{S} \sum_{s=1}^{S} [\hat                             y_s(\vx) - \E(\hat y)]^2.
\end{align}
This method is usually called MC dropout since the predictive output distribution is approximated by the Monte Carlo sampling. It can be proved that a neural network model with dropout mask for each layer can mathematically approximate a deep Gaussian process~\cite{gal2016dropout}. More details and theory properties on this MC dropout approach can be found in~\cite{gal2016uncertainty}.

\subsection{Gradient-Enhanced Surrogate Model}

In many real-world optimization problems, the gradient information is available with low or even trivial additional cost. For many machine learning applications, such as training a deep neural network, the gradient of the hyperparameters can be easily obtained or cheaply estimated by different methods~\cite{maclaurin2015gradient,fu2016drmad,wu2017bayesian,bengio2000gradient}. The continuous adjoint gradient for aerodynamic shape optimization problems can be approximated at low computational cost~\cite{jameson2003reduction}. Recently, some works show that utilizing gradient information is beneficial for multi-objective optimization algorithms~\cite{fliege2016method,wang2017hypervolume}.

In the expensive optimization setting, the gradient information is also useful for building a more accurate Gaussian process model with fewer evaluated solutions~\cite{solak2003derivative}, which might lead to more efficient surrogate-assisted optimization algorithms~\cite{ahmed2016we,wu2017exploiting}. However, the cost of incorporating gradient information into a Gaussian process with $N$ training samples in a $K$ dimensional decision space is $O(K^3N^3)$, which prevents its use for solving problems that have many decision variables or require a large number of evaluations.

Recently, a Sobolev training method~\cite{google2017sobolev} is proposed for incorporating available gradient information into the neural network training process. In this paper, we extend its use for building gradient-enhanced surrogate models to solve expensive optimization problems. The Sobolev training method does not change the neural network structure. It encodes the gradient information into neural network training by merely modifying the loss function as:

\begin{align}
L &=  L_{e}
   + L_{g} \nonumber \\
   &=\sum_{i=1}^{N} l(\hat f(\vx_i|\vW),f(\vx_i))
   + \sum_{i=1}^{N}  l (\nabla \hat f(\vx_i|\vW),\nabla f(\vx_i)) \nonumber \\
      &=\sum_{i=1}^{N} [\hat f(\vx_i|\vW) - f(\vx_i)]^2
   + \sum_{i=1}^{N}  [(\nabla \hat f(\vx_i|\vW),\nabla f(\vx_i))]^2
\end{align}
where $L_{e} = \sum_{i=1}^{N} l(\hat f(\vx_i|\vW),f(\vx_i))$ is the original error loss function and $L_{g}  = \sum_{i=1}^{N} l (\nabla \hat f(\vx_i|\vW),\nabla f(\vx_i))$ is the gradient loss. $\nabla \hat f(\vx_i|\vW)$ and $\nabla f(\vx_i)$ are the gradient vectors with respect to the Bayesian neural network output and the original objective function. In this paper, we want to build regression models to approximate the objective functions. Therefore, we use mean squared error (MSE) for both error loss and gradient loss. By training with this Sobolev loss function, a neural network model can encode not only the objective values but also the gradient information of the ground truth function, and hence can provide more accurate predictions for new solutions.

Combining this Sobolev training method along with the Bayesian neural network with practical MC dropout inference, we can build gradient-enhanced scalable surrogate models for large scale multi-objective optimization problems. The details of the model building will be discussed in section III.

\subsection{Batch Optimization and Parallel Evaluation}

Many real-world applications support parallel evaluation. For example, training multiple large deep neural networks with different structures and hyperparameters in parallel is a common practice for neural network structure search~\cite{zoph2016neural}. Batch function evaluation can significantly reduce the wall-clock time for the whole optimization process.

In multi-objective Bayesian optimization, the acquisition function $\alpha(\vx|D_t)$ is depended on all evaluated solutions. Although the goal is to obtain a set of Pareto optimal solutions, many surrogate-assisted algorithms use a scalar acquisition function to select only one single solution for evaluation at each iteration.  ParEGO ~\cite{knowles2006parego} randomly combines the multiple objectives into a single objective optimization problem at each iteration. SMS-EGO~\cite{ponweiser2008multiobjective} defines a scalar S-metric for searching the most promising solution for evaluation. These methods do not support parallel evaluation and batch optimization.

Some surrogate-assisted multi-objective optimization algorithms support parallel evaluation. For example, MOEA/D-EGO~\cite{zhang2010expensive} clusters all subproblems into a few groups and simultaneously evaluates the solutions with the highest expected improvements in each group. K-RVEA~\cite{chugh2016surrogate} uses adaptive reference vectors and cluster method to select multiple solutions for evaluation. However, these methods do not consider those already evaluated solutions for new solutions selection. This ignorance might lead to inferior batch selection and hence poorer optimization performance. In this paper, we propose an efficient selection strategy for batch multi-objective optimization, of which the details are discussed in section III.

\section{The Proposed Algorithm: BS-MOBO }

In this section, we propose a batched scalable multi-objective Bayesian optimization (BS-MOBO) algorithm for large scale expensive multi-objective optimization. This algorithm can easily incorporate available gradient information and supports batch optimization. The general algorithm framework is shown in \textbf{Algorithm}~\ref{alg_framewrk}.

\begin{algorithm}[h]
	\begin{algorithmic}[1]
    \STATE \textbf{Input:} MOP(1), Number of Iterations T
    \STATE Initialize the dataset $D_0$
	\FOR{$t = 1$ to $T$}
		\STATE Train scalable probabilistic model $\hat \vF(\vx|D_{t-1},\vW)$ with gradient information  \COMMENT{\textbf{Subsection III.A}}
    	\STATE Generate $k$ points for evaluation by optimizing the batch acquisition  $\alpha(\vX|D_{t-1},\vW)$ \COMMENT{\textbf{Subsection III.B}}
    	\STATE Evaluate the generated points, update dataset $D_{t}$
	\ENDFOR
    \STATE \textbf{Output:} nondominated solutions in $D_T$
	\end{algorithmic}
	\caption{BS-MOBO}
	\label{alg_framewrk}
\end{algorithm}

The main steps of this framework are:

\begin{itemize}
    \item \textbf{Initialization:} BS-MOBO initializes a dataset $D_0$ with a set of $N_I$ solutions using Latin hypercube sampling~\cite{mckay2000comparison}. All solutions are evaluated and will be used for model building.
    \item \textbf{Model Training:}  At each iteration $t$, total $m$ surrogate models will be built for approximating the objective functions based on the evaluated solutions in $D_{t-1}$, where $m$ is the number of objective functions. Details of the model training are discussed in \textbf{Subsection III.A}.
    \item \textbf{Solution Selection:} BS-MOBO selects $k$ promising solutions in batch for expensive parallel evaluation at each iteration via a two-stage selection process, of which the details are presented in \textbf{Subsection III.B}.
    \item \textbf{Update:} At the end of each iteration, the data set will be updated with the $k$ selected and evaluated solutions $D_t = D_{t-1} \cup \{(\vx_i,F(\vx_i))| i = 1,\cdots,k\}$ and will be used for updating the surrogate models.

\end{itemize}

\subsection{Scalable Surrogate Model with Gradient Information}

BS-MOBO uses a scalable Bayesian neural network with MC dropout inference as the surrogate model. With the Sobolev training technique, this model can easily incorporate gradient information into model training when it is available.

\subsubsection{Model Training}

In BS-MOBO, we build a fully connected neural network with two hidden layers as the surrogate model. The neural network has the same structure as the model in~\cite{google2017sobolev} for regression, of which the number of nodes for each hidden layer is $256$ and the activation function is ReLu. To do Bayesian inference by MC dropout, we use dropout in all hidden layers and the dropout rate is 0.05 for each node. Although the neural network structure and the dropout rate can be further optimized or adaptively adjusted for different problems, in this paper, we use a fixed structure to highlight its robustness and generality for solving various expensive multi-objective optimization problems.


In addition to its scalability, with Sobolev training, the neural network can efficiently incorporate available gradient information. As mentioned in section II, we do not have to change the network structure. The only thing we have to do is to consider the gradient information in the loss function. In this paper, we build independent surrogate models with the identical neural network structure for all objective functions. The algorithm of model training is shown in \textbf{Algorithm~\ref{alg_training}}.

\begin{algorithm}[h]
	\begin{algorithmic}[1]
    \STATE \textbf{Input:}  $D_t = \{(\vx_i,\vF(\vx_i))| i = 1,\cdots,N_{t}\}$ \\ (optional) gradient information $\{\nabla \vF(\vx_i)| i = 1,\cdots,N_{t}\}$
	\FOR{$j = 1$ to $m$}
		\STATE Initialize a neural network model $\hat f_j(\vx|\vW)$
        \IF{gradient information is available}
        	\STATE Train the model by optimizing the Sobolev loss
            \begin{align*}
            L = &\sum_{i=1}^{N_t} l(\hat f_j(\vx_i|\vW),F_j(\vx_i)) \\
               &+ \sum_{i=1}^{N_t}l(\nabla \hat f_j(\vx_i|\vW),\nabla F_j(\vx_i))
            \end{align*}
        \ELSIF{gradient information is not available}
        	\STATE Train the model by optimizing the error loss
            \begin{align*}
            L = \sum_{i=1}^{N_t} l(\hat f_j(\vx_i|\vW),F_j(\vx_i))
            \end{align*}
        \ENDIF
	\ENDFOR
    \STATE \textbf{Output:} $m$ trained models $\{\hat f_j(\vx|\vW)| j = 1,\cdots,m\}$
	\end{algorithmic}
	\caption{Model Training}
	\label{alg_training}
\end{algorithm}

\subsubsection{Model Prediction}

We use Bayesian inference with MC dropout approximation to obtain the predictive mean and variance for each objective function. For a new solution $\vx$, its predictive mean and variance can be obtained in \textbf{Algorithm~\ref{alg_prediction}}. The number of Monte Carlo sampling $S$ is a hyperparameter, and we set it to $20$ for all experiments conducted in this paper.

\begin{algorithm}[h]
	\begin{algorithmic}[1]
    \STATE \textbf{Input:}  a newly generated solution $\vx$ \\
            $m$ trained neural network models $\{\hat f_j(\vx|\vW)\}_{j=1}^{m}$
	\FOR{$j = 1$ to $m$}
    	\FOR{$s = 1$ to $S$}
    		\STATE Sample the weights $\widehat \vW_s$ for $\hat f_j(\vx|\widehat \vW_s)$ according to the dropout distribution
            \STATE Obtain the predictive value $\hat y_{js} = \hat f_j(\vx|\widehat \vW_s)$
		\ENDFOR
        \STATE Calculate the predictive mean and standard deviation
        \begin{align*}
           & \hat y_{j} = \frac{1}{S} \sum_{s=1}^{S} \hat y_{js}, \\
           & \hat \sigma_{j}^2 = \frac{1}{S} \sum_{s=1}^{S} [\hat y_{js} - \hat y_{j}]^2 
        \end{align*}
	\ENDFOR
    \STATE \textbf{Output:} the predictive mean vector $(\hat y_{1},...,\hat y_{m})$
    \STATE  the predictive standard deviation vector $(\hat \sigma_{1},...,\hat \sigma_{m})$
	\end{algorithmic}
	\caption{Prediction}
	\label{alg_prediction}
\end{algorithm}

\begin{figure}[h]
    \centering
    \includegraphics[width= 0.5\textwidth]{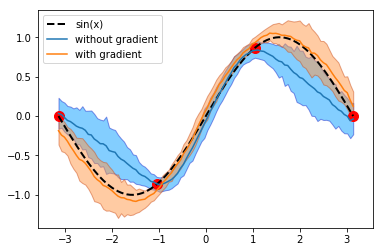}
    \caption{\label{fig:gradient} The predictive mean (solid line) $\pm$ two standard deviations (shade area) obtained by surrogate models with and without gradient information for sin(x).}
\end{figure}

Combining the Sobolev training and practical MC dropout inference, we can build a scalable surrogate model with a fewer number of evaluated solutions. As shown in Fig.~\ref{fig:gradient}, with only four evaluated solutions, the surrogate model with gradient can approximate the $sin(x)$ function very well. The optimal solution which has the maximum or minimum value can be better located using the surrogate model with gradient information.

\begin{figure}[H]
    \centering
    \includegraphics[width= 0.50\textwidth]{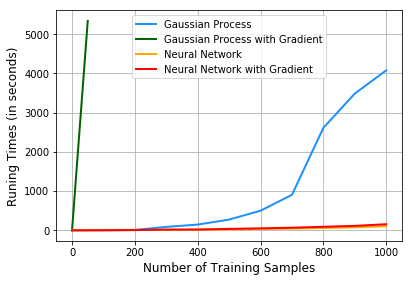}
    \caption{\label{fig:runtime} The training time of Gaussian process and Bayesian neural network with and without gradient information.}
\end{figure}

We also compare the training time of the Gaussian process and Bayesian neural network with and without gradient information for a function with $30$ decision variables in Fig.~\ref{fig:runtime}. It is obvious that the training time for building a Bayesian neural network with MC dropout increases very slowly when the number of training samples becomes larger. The extra computational cost for incorporating gradient information via Sobolev training is negligible, so the training time is almost coincident with the training time for training a model without gradient. This scalable ability makes it suitable for solving expensive optimization problems with larger decision space and requires more evaluation budget.

The scalable surrogate model is an important building block for the proposed scalable BS-MOBO algorithm. The comparisons between the proposed surrogate model and the Gaussian process in the surrogate-assisted expensive multi-objective optimization will be conducted and discussed in the experimental section.

\subsection{Batch Optimization}

Batch optimization is another key challenge for solving expensive multiobjective optimization problems, especially when the number of required function evaluations is large. As pointed out in the previous sections, the goal of BS-MOBO is to generate a set of promising solutions $D_T$ to provide a good trade-off for a given multiobjective optimization problem with a limited evaluation budget $T$. Therefore, at each iteration $t$, the batch acquisition function $\alpha(\vX|D_{t-1})$ should depend on all already-evaluated solutions $D_{t-1}$.

\subsubsection{Batch Acquisition Function}
The expected hypervolume improvement (EHI)~\cite{wagner2010expected} is a widely-used acquisition function for expensive multi-objective optimization problems. However, a standard EHI algorithm can only generate one promising solution for evaluation at each iteration~\cite{emmerich2008computation}. It is natural to generalize it to a batch form:
\begin{align}
&\textbf{B-EHI}(\vX|D_t) = \textbf{B-EHI}((\vx_{t+1}^{(1)},...,\vx_{t+1}^{(k)})|D_t)  \\
&= \E[\vH(\vF(D_t) \cup \{\vF (\vx_{t+1}^{(1)}),...,\vF(\vx_{t+1}^{(k)})\}) - \vH(\vF(D_t))],  \nonumber
\end{align}
where $\vH(\vY)$ is the hypervolume of all points in the set $\vY$ and $\vF(D_t)$ is the set of all objective values in $D_t$. By optimizing the batch expected hypervolume improvement acquisition function $\textbf{B-EHI}(\vX|D_t)$, we can obtain a set of $k$ promising solutions $\vX$ for parallel expensive evaluations. However, directly optimizing the above batch acquisition is very difficult, since we do not know how to directly find those $k$ promising solutions under the expectation with distribution $p(\vF (\vx_{t+1}^{(1)}),...,\vF(\vx_{t+1}^{(m)})|D_t)$.

Inspired by the upper confidence bound (UCB) acquisition function~\cite{Srinivas10} for single objective Bayesian optimization, we propose a batch hypervolume upper confidence bound (B-HUCB) acquisition for batch multi-objective Bayesian optimization:
\begin{align}
&\textbf{B-HUCB}(\vX|D_t) = \textbf{B-HUCB}((\vx_{t+1}^{(1)},...,\vx_{t+1}^{(k)})|D_t) \nonumber \\
&= \vH(\vF(D_t) \cup \{\vG (\vx_{t+1}^{(1)}),...,\vG(\vx_{t+1}^{(k)})\}) - \vH(D_t),
\end{align}
where $\boldsymbol{G(x)} = (\hat{\mu}_1(\vx) - \hat{\sigma}_1(\vx), \ldots, \hat{\mu}_m(\vx) - \hat{\sigma}_m(\vx))$ is the vector of lower confidence bound (LCB) for all objective functions of a minimization MOP. This batch acquisition function measures the hypervolume improvement by the LCB of a set of solutions $\vG(\vX) = \{\vG (\vx_{t+1}^{(1)}),...,\vG(\vx_{t+1}^{(k)})\}$ with respect to the set of all already evaluated solutions $D_t$.

The proposed batch acquisition function avoids the calculation of the batch expected hypervolume improvement. It also provides a good exploitation-exploration trade-off to select a set of promising solutions for expensive evaluations by considering all already-evaluated solutions $D_t$. However, it is still difficult to directly find a set of $k$ solutions $\vX_k$ to maximize $\textbf{B-HUCB}(\vX|D_t)$. In BS-MOBO, we propose a two-step algorithm for the solution selection process.

\subsubsection{Two-Step Batch Solutions Selection}

The algorithm framework of the proposed solutions selection process is shown in Algorithm~\ref{alg_selection}.
\begin{algorithm}[H]%
        	\caption{Two-Step Solutions Selection}
       	 	\label{alg_selection}
	\begin{algorithmic}[1]
	\STATE Find a set of $p$ candidate solutions $\vX_p$ by solving the surrogate multi-objective problem
     \STATE Select $k$ solutions among those $p$ candidates by optimizing
     \[\vX_k  = \arg \max_{\vX \subset \vX_p} \textbf{B-HUCB}(\vX|D_t)
     \]
     \STATE Evaluate all $k$ solutions $\vX_k$ in parallel
	\end{algorithmic}
\end{algorithm}
At the first step, based on the surrogate models, we can define a surrogate multi-objective optimization problem with respect to the lower confidence bound vector:
\begin{eqnarray}
    \label{surrogate_mop}
    \begin{aligned}
        &\min~\boldsymbol{G(x)} = (g_1(\boldsymbol{x}), \ldots, g_m(\boldsymbol{x})) \in \mathbb{R}^m\\
        &s.t.~~~ \textbf{x} \in \Omega \subset \mathbb{R}^n
    \end{aligned}
\end{eqnarray}
where $g_i(\vx) = \hat{\mu}_i(\vx) - \hat{\sigma}_i(\vx)$ is the lower confidence bound of objective function $f_i(\vx)$ at solution $\vx$. By optimizing the surrogate problems, we can obtain an approximate Pareto set of solutions $\vX_p$.

At the second stage, we can select the optimal subset of solution $\vX_k \subset \vX_p$ for parallel evaluations by optimizing the batch hypervolume upper confidence bound acquisition $\textbf{B-HUCB}(\vX|D_t)$.

In the proposed algorithm, we use MOEA/D to solve the surrogate multi-objective optimization problem. With Tcheycheff decomposition, the surrogate MOP is decomposed into a set of scalar surrogate subproblems:

\begin{equation}
\begin{aligned}
             &\text{minimize \quad  } h(\vx|\lambda,z^*) = \max_{1 \leq i \leq m} \{\lambda_i \abs{g_i(\vx) - z^*_i}\} \\
             &\text{subject to  \quad } \vx \in \Omega
\end{aligned}
\end{equation}
where $\lambda = (\lambda_1,...,\lambda_m)$ is a weight vector, i.e., $\lambda_i \geq 0$ for all $i = 1,...,m $ and $\sum_{i=1}^{m}{\lambda_i = 1}$. In addition, $z^* = (z^*_1,...,z^*_m)$ is the reference point, where $z^*_i < \min\{g_i(\vx)| \vx \in \Omega\}$ for each $i = 1,...,m$. We solve this surrogate MOP with population size $p$. The obtained $p$ solutions $\vX_p$ will be used as the candidate pool for the second step.

\begin{figure}[H]
\centering
\includegraphics[width = 0.5\textwidth]{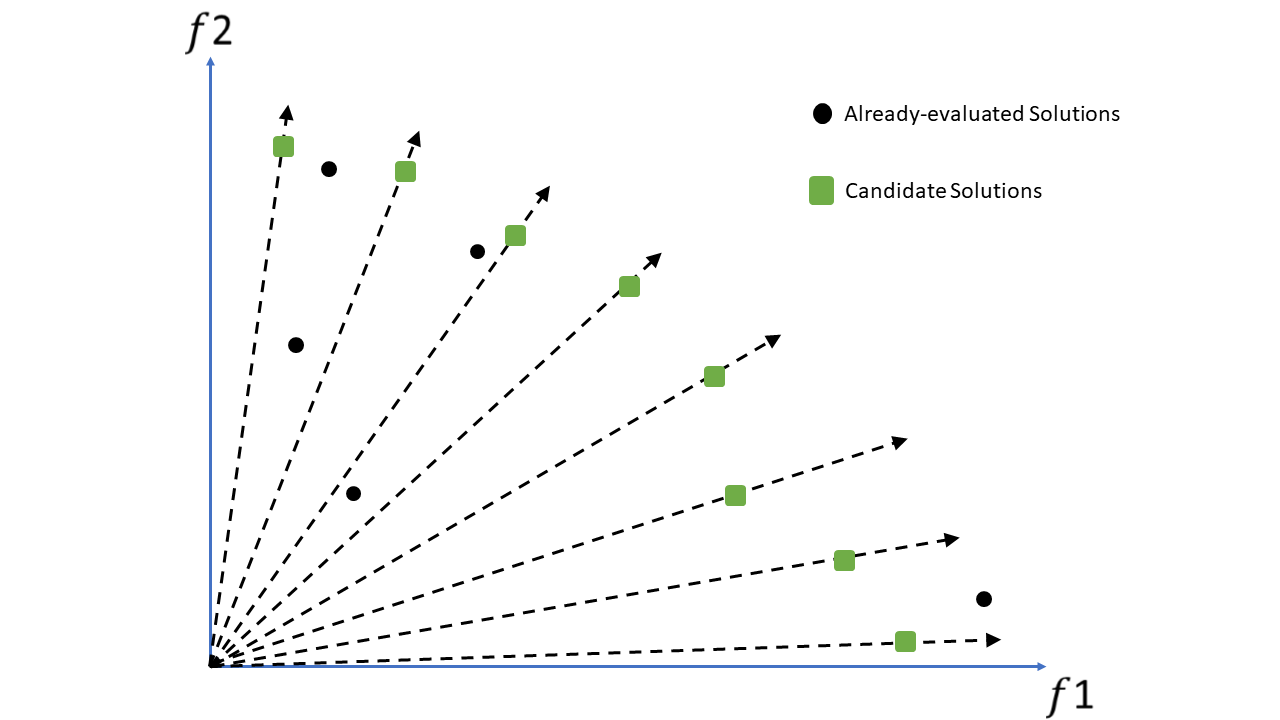}\\
\includegraphics[width = 0.5\textwidth]{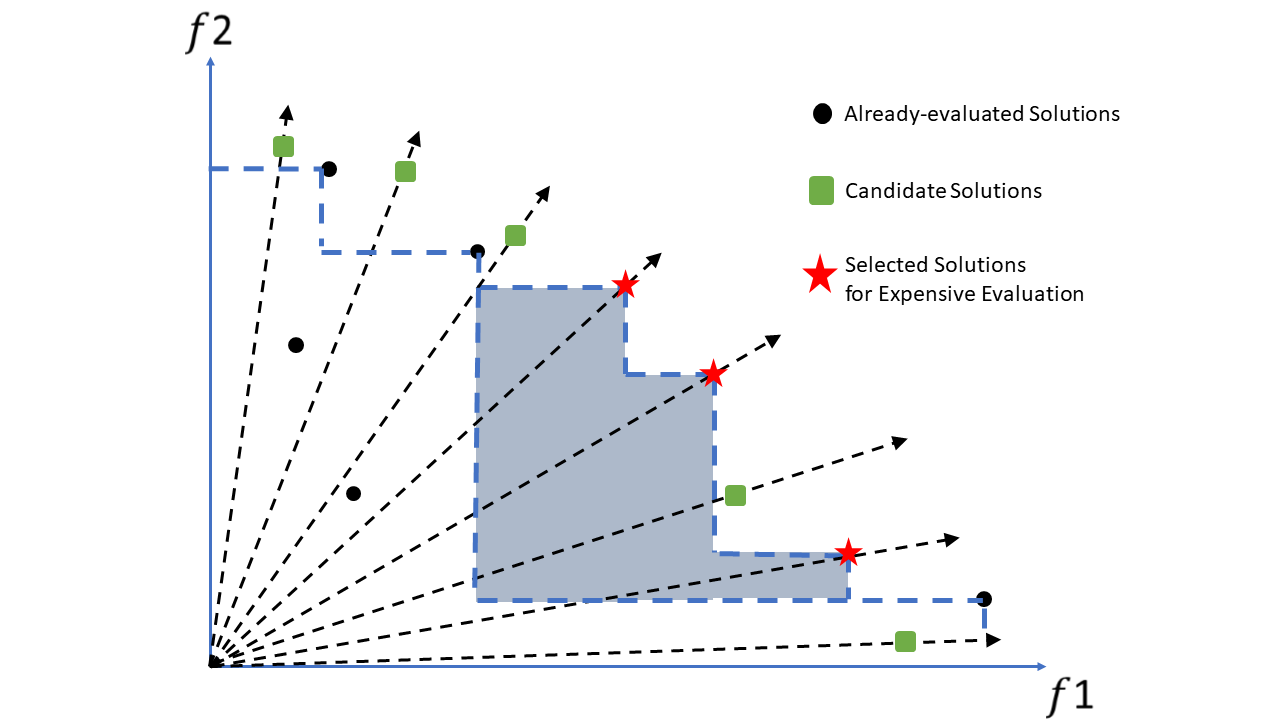}
\caption{\label{Solutions_Selection} An illustration of the two-step solutions selection process.}
\end{figure}

Once have obtained $p$ candidate solutions, the next step is to select $k$ solutions for evaluations. If $p = k$, we just evaluate all $p$ candidates. However, in BS-MOBO, it is more often that $p > k$ since: 1) the available batch size for many real-world application is usually limited (e.g., k = 10) and we need a larger number of solutions to approximate the surrogate Pareto front (e.g., p = 100); and 2) by choosing $k$ solutions from $p$ candidates, we can better select a set of promising solutions to maximize $\textbf{B-HUCB}(\vX|D_t)$ as shown in Fig.~\ref{Solutions_Selection}.

To choose the best $k$ candidate solutions for expensive evaluation, we have to take all already-evaluated solutions into consideration. It is still difficult since we have to find the best subset of $k$ solutions among all $C_{p,k}$ combinations. This hypervolume subset selection problem itself is computational expensive~\cite{bader2011hype}. In BS-MOBO, we use a greedy approach to select the $k$ candidate solutions as shown in \textbf{Algorithm}~\ref{alg_final_selection}.  In this algorithm, we first initialize a value set $\vV$ to store the objective values of all already-evaluated solutions in $D_t$. At each iteration, by assuming that the objective values of all candidate solutions in $\vX_p$ are equal to their lower confidence bound $\vG(\vx)$, we select the most promising solution in the candidate pool $\vX_p$ which maximizes the hypervolume contribution with respect to the value set $\vV$. The selected solution will be moved from the candidate pool $\vX_p$ to the selected set $\vX_k$ and its lower confidence bound value will be added into the value set $\vV$. After $k$ iteration, total $k$ promising solutions in $\vX_k$ are selected for expensive parallel evaluations.

\begin{algorithm}[h]
	\begin{algorithmic}[1]
    \STATE \textbf{Input:} Solutions in the candidate pool $\vX_p$, \\
            All already-evaluated solutions $D_t$
    \STATE Initialize the set of selected solutions $\vX_k = \emptyset$
    \STATE Initialize the value set $\vV = \vF(D_t)$
	\FOR{$i = 1$ to $k$}
		\STATE Select $\vx = \arg\max_{\vx \in X_p} [\vH(\vV \cup \{\vG (\vx)\}) - \vH(\vV)]$
    	\STATE Update $\vX_k = \vX_k \cup \{\vx\}$ and $\vX_p = \vX_p \setminus \{\vx\}$
        \STATE Update $\vV = \vV \cup \{\vG(\vx)\}$
	\ENDFOR
    \STATE \textbf{Output:} Set of selected solutions $\vX_k$
	\end{algorithmic}
	\caption{Greedy Promising Solutions Selection}
	\label{alg_final_selection}
\end{algorithm}

\section{Experimental Studies}

In this section, we empirically validate the performance of the proposed BS-MOBO algorithm on various benchmark functions as well as on a real-world application problem.

\subsection{Experimental Results on Small Scale Benchmark Problems}

\begin{table*}[t]
\centering
\setlength\tabcolsep{1.5pt}
\caption{Means[Stds](Rank) of IGD Values on 8 Dimensional Problems Obtained by Different Algorithms with 160 Function Evaluations.}
\label{results_table_small}
\begin{tabular}{c|ccc|cc|ccc}
\hline
\multicolumn{1}{l|}{}                                & \multicolumn{3}{c|}{Sequential Evaluation}                                                          & \multicolumn{2}{c|}{Batch Evaluation (batch size = 5)}                       & \multicolumn{3}{c}{Ours (batch size = 5)}                                                                                                                           \\
\multicolumn{1}{l|}{\multirow{-2}{*}{Test Instance}} & ParEGO               & HypI                 & SMS-EGO                                               & K-RVEA                                                & MOEA/D-EGO           & BS-MOBO-GP                                             & BS-MOBO                                                & BS-MOBO\_g                                            \\ \hline
ZDT1                                                 & 0.186{[}0.019{]}(8)+ & 0.080{[}0.013{]}(7)+ & 0.027{[}0.003{]}(5)+                                  & 0.054{[}0.012{]}(6)+                                  & 0.023{[}0.007{]}(4)+ & \cellcolor[HTML]{C0C0C0}\textbf{0.008{[}0.003{]}(1)=} & 0.017{[}0.002{]}(3)+                                  & 0.010{[}0.004{]}(2)                                  \\
ZDT2                                                 & 0.270{[}0.012{]}(8)+ & 0.176{[}0.024{]}(7)+ & 0.124{[}0.017{]}(6)+                                  & 0.031{[}0.006{]}(5)+                                  & 0.028{[}0.010{]}(4)+ & 0.015{[}0.003{]}(2)=                                  & 0.020{[}0.004{]}(3)+                                  & \cellcolor[HTML]{C0C0C0}\textbf{0.013{[}0.004{]}(1)} \\
ZDT3                                                 & 0.169{[}0.046{]}(7)+ & 0.166{[}0.032{]}(6)+ & 0.031{[}0.012{]}(2)-                                  & 0.171{[}0.050{]}(8)+                                  & 0.092{[}0.011{]}(4)+ & \cellcolor[HTML]{C0C0C0}\textbf{0.020{[}0.006{]}(1)-} & 0.133{[}0.041{]}(5)+                                  & 0.078{[}0.017{]}(3)                                  \\
ZDT4                                                 & 46.55{[}1.544{]}(6)+ & 50.64{[}2.231{]}(7)+ & 44.85{[}1.247{]}(4)+                                  & 27.80{[}7.432{]}(2)+                                  & 57.87{[}1.431{]}(8)+ & 45.09{[}12.83{]}(5)+                                  & 39.78{[}8.121{]}(3)+                                 & \cellcolor[HTML]{C0C0C0}\textbf{21.47{[}8.126{]}(1)} \\
ZDT6                                                 & 0.605{[}0.110{]}(7)+ & 0.529{[}0.117{]}(5)+ & 0.226{[}0.037{]}(2)-                                  & 0.970{[}0.320{]}(8)+                                  & 0.441{[}0.041{]}(4)+ & \cellcolor[HTML]{C0C0C0}\textbf{0.179{[}0.014{]}(1)-} & 0.582{[}0.139{]}(6)+                                  & 0.382{[}0.030{]}(3)                                  \\
DTLZ1                                                & 140.6{[}12.84{]}(8)+ & 87.78{[}9.241{]}(2)= & 139.2{[}21.82{]}(7)+                                  & \cellcolor[HTML]{C0C0C0}\textbf{83.37{[}19.70{]}(1)=} & 116.3{[}11.09{]}(5)+ & 93.16{[}5.871{]}(4)+                                   & 121.1{[}17.82{]}(6)+                                  & 89.87{[}9.840{]}(3)                                  \\
DTLZ2                                                & 0.232{[}0.024{]}(4)+ & 0.444{[}0.131{]}(8)+ & 0.339{[}0.078{]}(6)+                                  & 0.148{[}0.006{]}(2)+                                  & 0.342{[}0.117{]}(7)+ & 0.326{[}0.108{]}(5)+                                  & 0.205{[}0.092{]}(3)+                                  & \cellcolor[HTML]{C0C0C0}\textbf{0.123{[}0.024{]}(1)} \\
F1                                                   & 0.035{[}0.010{]}(4)+ & 0.082{[}0.017{]}(7)+ & 0.047{[}0.019{]}(5)+                                  & 0.102{[}0.005{]}(8)+                                  & 0.062{[}0.012{]}(6)+ & \cellcolor[HTML]{C0C0C0}\textbf{0.009{[}0.003{]}(1)-} & 0.031{[}0.018{]}(3)=                                  & 0.028{[}0.011{]}(2)                                  \\
F2                                                   & 0.050{[}0.012{]}(5)+ & 0.104{[}0.017{]}(7)+ & 0.028{[}0.007{]}(3)+                                  & 0.278{[}0.009{]}(8)+                                  & 0.083{[}0.019{]}(6)+ & \cellcolor[HTML]{C0C0C0}\textbf{0.008{[}0.002{]}(1)-} & 0.048{[}0.024{]}(4)+                                  & 0.021{[}0.009{]}(2)                                  \\
F3                                                   & 0.780{[}0.134{]}(3)+ & 2.036{[}0.278{]}(8)+ & 0.952{[}0.171{]}(4)+                                  & 1.406{[}0.231{]}(7)+                                  & 1.341{[}0.191{]}(6)+ & \cellcolor[HTML]{C0C0C0}\textbf{0.490{[}0.061{]}(1)-} & 1.310{[}0.276{]}(5)+                                  & 0.682{[}0.219{]}(2)                                  \\
F4                                                   & 0.186{[}0.047{]}(6)+ & 0.279{[}0.141{]}(8)+ & \cellcolor[HTML]{C0C0C0}\textbf{0.125{[}0.071{]}(1)=} & 0.205{[}0.024{]}(7)+                                  & 0.176{[}0.089{]}(5)+ & 0.145{[}0.078{]}(4)+                                  & 0.129{[}0.044{]}(2)=                                  & 0.134{[}0.032{]}(3)                                  \\
F5                                                   & 0.102{[}0.013{]}(3)- & 0.157{[}0.034{]}(4)- & 0.076{[}0.025{]}(2)-                                  & 0.361{[}0.044{]}(8)+                                  & 0.234{[}0.031{]}(5)- & \cellcolor[HTML]{C0C0C0}\textbf{0.046{[}0.037{]}(1)-} & 0.293{[}0.054{]}(7)+                                  & 0.268{[}0.061{]}(6)                                  \\
F6                                                   & 0.211{[}0.090{]}(3)+ & 0.283{[}0.124{]}(6)+ & 0.326{[}0.098{]}(7)+                                  & 0.341{[}0.014{]}(8)+                                  & 0.272{[}0.071{]}(5)+ & \cellcolor[HTML]{C0C0C0}\textbf{0.112{[}0.062{]}(1)-} & 0.252{[}0.045{]}(4)+                                  & 0.196{[}0.057{]}(2)                                  \\
F7                                                   & 0.393{[}0.083{]}(3)= & 0.583{[}0.121{]}(7)+ & 0.421{[}0.091{]}(4)+                                  & 0.887{[}0.133{]}(8)+                                  & 0.581{[}0.137{]}(6)+ & \cellcolor[HTML]{C0C0C0}\textbf{0.376{[}0.075{]}(1)=} & 0.502{[}0.024{]}(5)+                                  & 0.382{[}0.034{]}(2)                                  \\
F8                                                   & 0.207{[}0.041{]}(2)= & 0.276{[}0.074{]}(7)+ & 0.230{[}0.078{]}(5)+                                  & 0.285{[}0.013{]}(8)+                                  & 0.248{[}0.051{]}(6)+ & 0.227{[}0.047{]}(4)+                                  & \cellcolor[HTML]{C0C0C0}\textbf{0.131{[}0.021{]}(1)-} & 0.221{[}0.037{]}(3)                                  \\
F9                                                   & 0.430{[}0.085{]}(5)+ & 0.476{[}0.112{]}(7)+ & 0.474{[}0.067{]}(6)+                                  & 0.504{[}0.151{]}(8)+                                  & 0.412{[}0.049{]}(4)+ & \cellcolor[HTML]{C0C0C0}\textbf{0.312{[}0.037{]}(1)-} & 0.399{[}0.022{]}(3)+                                  & 0.377{[}0.014{]}(2)                                  \\
F10                                                  & 8.982{[}1.241{]}(4)+ & 135.3{[}7.824{]}(7)+ & 55.64{[}17.29{]}(6)+                                  & 136.8{[}6.261{]}(8)+                                  & 43.28{[}15.56{]}(5)+ & 0.822{[}0.022{]}(3)=                                  & 0.815{[}0.012{]}(2)=                                  & \cellcolor[HTML]{C0C0C0}\textbf{0.809{[}0.008{]}(1)} \\
UF1                                                  & 0.175{[}0.013{]}(4)- & 0.239{[}0.021{]}(6)- & 0.155{[}0.010{]}(2)-                                  & 0.181{[}0.012{]}(5)-                                  & 0.162{[}0.014{]}(3)- & \cellcolor[HTML]{C0C0C0}\textbf{0.059{[}0.015{]}(1)-} & 0.635{[}0.154{]}(8)+                                  & 0.573{[}0.102{]}(7)                                  \\
UF2                                                  & 0.205{[}0.020{]}(8)+ & 0.192{[}0.017{]}(7)+ & 0.133{[}0.031{]}(5)+                                  & 0.138{[}0.021{]}(4)+                                  & 0.171{[}0.014{]}(6)+ & 0.121{[}0.019{]}(2)+                                  & 0.131{[}0.028{]}(3)+                                  & \cellcolor[HTML]{C0C0C0}\textbf{0.107{[}0.014{]}(1)} \\
UF3                                                  & 1.478{[}0.055{]}(8)+ & 1.438{[}0.074{]}(7)+ & 0.821{[}0.145{]}(2)+                                  & 0.887{[}0.202{]}(3)+                                  & 1.417{[}0.215{]}(6)+ & \cellcolor[HTML]{C0C0C0}\textbf{0.701{[}0.140{]}(1)-} & 1.385{[}0.162{]}(5)+                                  & 1.212{[}0.097{]}(4)                                  \\
UF4                                                  & 0.110{[}0.015{]}(7)+ & 0.131{[}0.017{]}(8)+ & 0.095{[}0.010{]}(3)=                                  & 0.101{[}0.012{]}(4)+                                  & 0.104{[}0.011{]}(5)+ & \cellcolor[HTML]{C0C0C0}\textbf{0.069{[}0.002{]}(1)=} & 0.108{[}0.012{]}(6)+                                  & 0.093{[}0.015{]}(2)                                  \\
UF5                                                  & 2.002{[}0.419{]}(4)- & 1.885{[}0.228{]}(3)- & \cellcolor[HTML]{C0C0C0}\textbf{1.619{[}0.192{]}(1)-} & 1.682{[}0.839{]}(2)-                                  & 2.651{[}0.219{]}(7)= & 2.286{[}0.175{]}(5)=                                  & 2.788{[}0.123{]}(8)=                                  & 2.507{[}0.388{]}(6)                                  \\
UF6                                                  & 1.961{[}0.181{]}(6)= & 1.746{[}0.241{]}(3)- & \cellcolor[HTML]{C0C0C0}\textbf{1.375{[}0.183{]}(1)-} & 1.562{[}0.243{]}(2)-                                  & 1.891{[}0.262{]}(5)- & 1.860{[}0.547{]}(4)-                                  & 2.807{[}0.230{]}(8)+                                  & 2.443{[}0.366{]}(7)                                  \\
UF7                                                  & 0.234{[}0.055{]}(3)- & 0.172{[}0.024{]}(2)- & 0.276{[}0.072{]}(4)-                                  & 0.287{[}0.114{]}(5)-                                  & 0.421{[}0.031{]}(7)= & \cellcolor[HTML]{C0C0C0}\textbf{0.125{[}0.018{]}(1)-} & 0.572{[}0.122{]}(8)+                                  & 0.411{[}0.016{]}(6)                                  \\
UF8                                                  & 0.346{[}0.174{]}(7)+ & 0.379{[}0.076{]}(8)+ & 0.312{[}0.102{]}(4)+                                  & 0.343{[}0.033{]}(6)+                                  & 0.297{[}0.143{]}(3)= & \cellcolor[HTML]{C0C0C0}\textbf{0.151{[}0.061{]}(1)-} & 0.326{[}0.091{]}(5)+                                  & 0.287{[}0.121{]}(2)                                  \\
UF9                                                  & 0.491{[}0.036{]}(8)+ & 0.453{[}0.103{]}(7)+ & 0.430{[}0.132{]}(6)+                                  & 0.341{[}0.036{]}(3)+                                  & 0.424{[}0.122{]}(5)+ & 0.279{[}0.035{]}(2)+                                  & 0.413{[}0.042{]}(4)+                                  & \cellcolor[HTML]{C0C0C0}\textbf{0.226{[}0.078{]}(1)} \\
UF10                                                 & 2.993{[}0.112{]}(4)+ & 7.817{[}0.385{]}(8)+ & 2.442{[}0.113{]}(2)+                                  & 3.606{[}0.007{]}(7)+                                  & 3.479{[}0.179{]}(6)+ & 3.123{[}0.134{]}(5)+                                  & 2.558{[}0.045{]}(3)+                                  & \cellcolor[HTML]{C0C0C0}\textbf{2.217{[}0.114{]}(1)} \\ \hline
+/=/-                                                & 20/3/4               & 21/1/5               & 18/2/7                                                & 22/1/4                                                & 21/3/3               & 8/6/13                                                & 22/4/1                                                & -                                                    \\
average rank                                         & 5.370                & 6.259                & 3.889                                                 & 5.593                                                 & 5.296                & 2.222                                                 & 4.555                                                 & 2.815                                                \\ \hline
\end{tabular}
\end{table*}

The BS-MOBO algorithm is originally proposed for solving large scale expensive multi-objective optimization problems with high dimensional decision space. But it can also be used for solving small scale problems. In this subsection, we first test its performance on problems with low dimensional decision space (8 decision variables). We compare BS-MOBO with some state-of-the-art Kriging-based surrogate-assisted MOEAs, such as ParEGO~\cite{knowles2006parego}, HypI~\cite{rahat2017alternative}, SMS-EGO~\cite{ponweiser2008multiobjective}, K-RVEA~\cite{chugh2016surrogate} and MOEA/D-EGO~\cite{zhang2010expensive} on four widely-used benchmark problem suits, namely, ZDT~\cite{zitzler1999multiobjective}, DTLZ~\cite{deb2002scalable}, UF~\cite{zhang2008multiobjective} and F problem suit in the RM-MEDA papers~\cite{zhang2008rm}.

It should be noticed that ParEGO, HypI and SMS-EGO only support sequential evaluation where only one solution can be evaluated at each iteration, while K-RVEA, MOEA/D-EGO and our proposed algorithm can evaluate solutions in batch. If two algorithms have similar performance with the same number of expensive evaluations, the one which supports batch evaluation is usually preferred for many real-world applications since it can significantly save the wall-clock time.

We use the Python code\footnote{http://bitbucket.org/arahat/gecco-2017} from~\cite{rahat2017alternative} to run the ParEGO, HypI and SMS-EGO. The K-RVEA implementation\footnote{http://bimk.ahu.edu.cn/index.php?s=/Index/Software/index.html} is available from the PlatEMO toolbox~\cite{tian2017platemo}. We implement the MOEA/D-EGO and the proposed algorithm BS-MOBO in Python\footnote{Will be open source once this paper is published.} where the neural networks are built with Pytorch\footnote{https://pytorch.org/}~\cite{paszke2017automatic}.

The parameters settings for this comparison are:

\begin{itemize}
    \item Dimension of decision space $n = 8$;
    \item Maximal number of evaluations $N_{FE} = 20n = 160$;
    \item Number of solutions in the initial dataset $N_I = 60$;
    \item Number of solutions to be evaluated in batch at each iteration for K-RVEA, MOEA/D-EGO, BS-MOBO and its variants $k = 5$, which means there are only 20 parallel evaluations ($(160 - 60)/5 = 20$) for these algorithms;
    \item Population size for K-RVEA, MOEA/D-EGO, BS-MOBO and its variants $p = 100$;
    \item Parameters for all compared algorithms are the same with their default settings reported in the literature;
\end{itemize}

We independently run each algorithm 25 times and compare their results using the inverted generational distance (IGD) metric and the hypervolume difference ($I_H^-$) metric~\cite{zitzler:2003}. To calculate the IGD and $I_H^-$ metric, we use 500 uniformly distributed Pareto-optimal points in PF for ZDT test instances and 990 points for DTLZ test instances as in~\cite{zhang:2007}, 500 and 990 points for two and three objective F test instances, 1000 points for UF1-7 and 10000 points for UF8-10 as in \cite{zhang2008multiobjective}. Due to the page limit, the experimental results of hypervolume difference are reported in the supplementary material.

To better study the effect of different components in the algorithm, we report the results of the proposed algorithm along with two different variants. BS-MOBO is the proposed algorithm with a Bayesian neural network surrogate model without gradient information. BS-MOBO\_g incorporates the gradient information into the surrogate model building via Sobolev training. To validate the effectiveness of the proposed batch optimization framework, BS-MOBO-GP replaces the Bayesian neural network with a standard Gaussian process model as the surrogate model.

Table~\ref{results_table_small} shows the mean IGD values, the standard deviations and the ranks of IGD means for all $27$ problems obtained by different algorithms.  The best result for each problem is highlighted. In addition, a Wilcoxon rank-sum test with significant level $0.05$ is conducted to compare the performances between BS-MOBO\_g and other algorithms, where the symbols $+/=/-$ indicate that BS-MOBO\_g performs significantly better than/equally with/significant worse than the compared algorithms respectively.

\subsubsection{The Effect of the Batch Optimization Framework}
It is obvious that BS-MOBO-GP achieves the best overall performance with average rank $2.222$ and rank the first for most test instances. Since BS-MOBO-GP uses the same Gaussian process surrogate model with other Kriging-based algorithms, this result confirms the advantage of our newly proposed batch optimization framework. BS-MOBO-GP is outperformed by other Kriging-based algorithms on the UF5 and UF6 test instances which both have disconnected Pareto set and many local Pareto optimal points. Our proposed batch selection method aims at finding a set of well-distributed solutions to improve the hypervolume at each iteration. But it seems that the convergence performance is poor and many solutions are got trapped by local Pareto points for these problems with disconnected Pareto fronts.

\subsubsection{Scalable Model with Very Few Solutions}
The performances of BS-MOBO are worse than those of BS-MOBO-GP for most test instances. The only difference between BS-MOBO and BS-MOBO-GP is the surrogate model (Bayesian neural network for BS-MOBO and Gaussian process for BS-MOBO-GP). These results are reasonable for the small scale problems with a very limited number of evaluated solutions. It is well known that Gaussian process modeling has excellent fitting performances for small datasets in low dimensional space~\cite{rasmussen2006gaussian}. On the contrary, neural networks usually need more training examples to achieve good fitting performance~\cite{goodfellow2016deep,lee2017deep}. However, the algorithm performance would be entirely different for large scale problems as can be seen in the next subsection. In addition, with a simple ensemble method, we can propose a hybrid algorithm which achieves good performance for both small and large scale problems. The details of the simple hybrid algorithm can be found in the supplementary material.

\subsubsection{The Effect of Gradient Information}
With available gradient information, BS-MOBO\_g significantly outperforms BS-MOBO on most test instances. It confirms that incorporating gradient information into the surrogate model can substantially improve the optimization algorithm's performance. It is interesting to observe that, with the help of gradient information, BS-MOBO\_g can obtain better or equally good performances with BS-MOBO-GP on many test instances even in the small scale setting. BS-MOBO\_g performs slightly worse than BS-MOBO on F4 and F8 test instances which both have three complicated periodic objective functions with nonlinear variable linkage.
For these problems, the Sobolev training method tries to match many local gradient directions with very few data points, and hence performs poorly to approximate the whole complicated function. Similar results have also been
reported in the Sobolov training paper~\cite{google2017sobolev}.

\begin{table*}[t]
\centering
\setlength\tabcolsep{1.5pt}
\caption{Means[Stds](Rank) of IGD Values on 50 Dimensional Problems Obtained by Different Algorithms with 1000 Function Evaluations.}
\label{results_table_large}
\begin{tabular}{c|cc|cc|ccc}
\hline
\multicolumn{1}{l|}{}                   & \multicolumn{2}{c|}{Model-Free Method}                                        & \multicolumn{2}{l|}{Batch Evaluation (batch size = 25)}                       & \multicolumn{3}{c}{Ours (batch size = 25)}                                                                                                                           \\
\multicolumn{1}{l|}{\multirow{-2}{*}{}} & MOEA/D               & HIGA-MO                                                & K-RVEA                                                & MOEA/D-EGO            & BS-MOBO-GP                                            & BS-MOBO                                               & BS-MOBO\_g                                            \\ \hline
ZDT1                                    & 1.197{[}0.107{]}(7)+ & 0.891{[}0.099{]}(6)+                                   & 0.505{[}0.430{]}(5)+                                  & 0.283{[}0.032{]}(4)+  & 0.078{[}0.012{]}(3)+                                  & 0.036{[}0.007{]}(2)=                                  & \cellcolor[HTML]{C0C0C0}\textbf{0.019{[}0.008{]}(1)}  \\
ZDT2                                    & 3.360{[}0.105{]}(7)+ & 2.367{[}0.102{]}(6)+                                   & 1.498{[}0.490{]}(5)+                                  & 0.588{[}0.113{]}(4)+  & 0.034{[}0.009{]}(3)+                                  & 0.028{[}0.012{]}(2)+                                  & \cellcolor[HTML]{C0C0C0}\textbf{0.021{[}0.006{]}(1)}  \\
ZDT3                                    & 1.665{[}0.158{]}(6)+ & 1.978{[}0.191{]}(7)+                                   & 0.487{[}0.134{]}(4)+                                  & 0.513{[}0.082{]}(5)+  & 0.165{[}0.112{]}(2)+                                  & 0.254{[}0.043{]}(3)+                                  & \cellcolor[HTML]{C0C0C0}\textbf{0.055{[}0.013{]}(1)}  \\
ZDT4                                    & 395.7{[}6.230{]}(2)- & \cellcolor[HTML]{C0C0C0}\textbf{292.5{[}25.864{]}(1)-} & 461.4{[}63.90{]}(3)-                                  & 673.93{[}77.91{]}(5)= & 686.4{[}42.98{]}(7)=                                  & 685.4{[}24.64{]}(6)=                                  & 654.87{[}11.21{]}(4)                                  \\
ZDT6                                    & 6.986{[}0.225{]}(6)+ & 7.094{[}0.147{]}(7)+                                   & 4.261{[}0.973{]}(5)+                                  & 3.962{[}0.334{]}(4)+  & \cellcolor[HTML]{C0C0C0}\textbf{0.554{[}0.192{]}(1)-} & 3.633{[}0.286{]}(3)+                                  & 2.343{[}0.241{]}(2)                                   \\
DTLZ1                                   & 1576{[}172.8{]}(5)+  & NA                                                     & 2166{[}232.4{]}(6)+                                   & 1430{[}246.8{]}(4)+   & 1232{[}182.6{]}(2)+                                   & 1272{[}14.21{]}(3)+                                   & \cellcolor[HTML]{C0C0C0}\textbf{1042.3{[}68.91{]}(1)} \\
DTLZ2                                   & 0.975{[}0.071{]}(4)+ & NA                                                     & 1.872{[}0.231{]}(5)+                                  & 1.983{[}0.413{]}(6)+  & 0.742{[}0.241{]}(3)+                                  & 0.348{[}0.083{]}(2)+                                  & \cellcolor[HTML]{C0C0C0}\textbf{0.244{[}0.081{]}(1)}  \\
F1                                      & 0.406{[}0.026{]}(7)+ & 0.294{[}0.009{]}(3)+                                   & 0.382{[}0.088{]}(5)+                                  & 0.374{[}0.034{]}(4)+  & 0.396{[}0.093{]}(6)+                                  & 0.264{[}0.025{]}(2)+                                  & \cellcolor[HTML]{C0C0C0}\textbf{0.219{[}0.043{]}(1)}  \\
F2                                      & 0.670{[}0.022{]}(6)+ & 0.525{[}0.013{]}(3)+                                   & 0.621{[}0.092{]}(5)+                                  & 0.683{[}0.033{]}(7)+  & 0.617{[}0.073{]}(4)+                                  & 0.423{[}0.012{]}(2)+                                  & \cellcolor[HTML]{C0C0C0}\textbf{0.343{[}0.088{]}(1)}  \\
F3                                      & 7.097{[}0.274{]}(7)+ & 6.308{[}0.123{]}(5)+                                   & 5.741{[}0.641{]}(4)+                                  & 6.798{[}0.241{]}(6)+  & 5.134{[}0.439{]}(3)+                                  & 3.872{[}0.413{]}(2)=                                  & \cellcolor[HTML]{C0C0C0}\textbf{3.640{[}0.143{]}(1)}  \\
F4                                      & 2.591{[}0.232{]}(3)+ & NA                                                     & 2.768{[}0.131{]}(6)+                                  & 2.634{[}0.182{]}(4)+  & 2.726{[}0.176{]}(5)+                                  & \cellcolor[HTML]{C0C0C0}\textbf{0.588{[}0.068{]}(1)=} & 0.627{[}0.083{]}(2)                                   \\
F5                                      & 0.546{[}0.042{]}(6)+ & 0.507{[}0.013{]}(4)+                                   & 0.562{[}0.092{]}(7)+                                  & 0.543{[}0.043{]}(5)+  & 0.504{[}0.062{]}(3)+                                  & 0.484{[}0.043{]}(2)+                                  & \cellcolor[HTML]{C0C0C0}\textbf{0.396{[}0.061{]}(1)}  \\
F6                                      & 0.604{[}0.098{]}(5)+ & 0.483{[}0.020{]}(2)+                                   & 0.663{[}0.132{]}(7)+                                  & 0.613{[}0.124{]}(6)+  & 0.505{[}0.094{]}(3)+                                  & 0.587{[}0.032{]}(4)+                                  & \cellcolor[HTML]{C0C0C0}\textbf{0.373{[}0.049{]}(1)}  \\
F7                                      & 6.696{[}0.112{]}(7)+ & 5.970{[}0.266{]}(3)+                                   & 5.421{[}0.341{]}(4)+                                  & 6.193{[}0.682{]}(5)+  & 6.382{[}0.563{]}(6)+                                  & 3.863{[}0.142{]}(2)+                                  & \cellcolor[HTML]{C0C0C0}\textbf{3.127{[}0.032{]}(1)}  \\
F8                                      & 1.907{[}0.461{]}(3)+ & NA                                                     & 2.317{[}0.172{]}(4)+                                  & 2.934{[}0.182{]}(6)+  & 2.809{[}0.198{]}(5)+                                  & 1.183{[}0.224{]}(2)+                                  & \cellcolor[HTML]{C0C0C0}\textbf{0.654{[}0.113{]}(1)}  \\
F9                                      & 6.103{[}0.432{]}(7)+ & 5.987{[}0.899{]}(6)+                                   & 2.829{[}0.542{]}(4)+                                  & 3.482{[}0.583{]}(5)+  & 0.783{[}0.143{]}(3)+                                  & \cellcolor[HTML]{C0C0C0}\textbf{0.262{[}0.043{]}(1)=} & 0.282{[}0.088{]}(2)                                   \\
F10                                     & 12644{[}876{]}(6)+   & 20004{[}972{]}(7)+                                     & 2799{[}541.2{]}(5)+                                   & 1783{[}342.9{]}(4)+   & 1341{[}231.8{]}(3)+                                   & 0.820{[}0.003{]}(2)=                                  & \cellcolor[HTML]{C0C0C0}\textbf{0.818{[}0.007{]}(1)}  \\
UF1                                     & 1.281{[}0.072{]}(4)+ & 1.382{[}0.082{]}(5)+                                   & 1.398{[}0.033{]}(6)+                                  & 1.124{[}0.083{]}(3)=  & 1.419{[}0.083{]}(7)+                                  & 1.185{[}0.032{]}(2)=                                  & \cellcolor[HTML]{C0C0C0}\textbf{1.104{[}0.033{]}(1)}  \\
UF2                                     & 0.429{[}0.138{]}(5)+ & 0.466{[}0.026{]}(6)+                                   & 0.342{[}0.076{]}(4)+                                  & 0.558{[}0.043{]}(7)+  & 0.321{[}0.113{]}(3)+                                  & \cellcolor[HTML]{C0C0C0}\textbf{0.121{[}0.011{]}(1)=} & 0.124{[}0.043{]}(2)                                   \\
UF3                                     & 0.851{[}0.051{]}(4)+ & 0.772{[}0.035{]}(3)+                                   & 0.976{[}0.087{]}(6)+                                  & 1.022{[}0.043{]}(7)+  & 0.950{[}0.042{]}(5)+                                  & 0.483{[}0.027{]}(2)+                                  & \cellcolor[HTML]{C0C0C0}\textbf{0.423{[}0.043{]}(1)}  \\
UF4                                     & 0.183{[}0.011{]}(5)+ & 0.203{[}0.003{]}(7)+                                   & 0.181{[}0.024{]}(4)+                                  & 0.187{[}0.018{]}(6)+  & \cellcolor[HTML]{C0C0C0}\textbf{0.114{[}0.013{]}(1)-} & 0.145{[}0.013{]}(3)+                                  & 0.124{[}0.043{]}(2)                                   \\
UF5                                     & 5.103{[}0.033{]}(3)+ & 5.729{[}0.344{]}(6)+                                   & 5.623{[}0.133{]}(7)+                                  & 5.472{[}0.048{]}(5)+  & 5.271{[}0.094{]}(4)+                                  & 5.054{[}0.077{]}(2)+                                  & \cellcolor[HTML]{C0C0C0}\textbf{4.383{[}0.034{]}(1)}  \\
UF6                                     & 4.458{[}0.046{]}(3)+ & 5.882{[}0.329{]}(5)+                                   & 6.242{[}0.072{]}(7)+                                  & 5.982{[}0.094{]}(6)+  & 5.241{[}0.143{]}(4)+                                  & \cellcolor[HTML]{C0C0C0}\textbf{3.228{[}0.133{]}(1)-} & 3.539{[}0.172{]}(2)                                   \\
UF7                                     & 1.668{[}0.231{]}(7)+ & 1.426{[}0.066{]}(4)+                                   & 0.982{[}0.124{]}(3)=                                  & 1.483{[}0.083{]}(5)+  & 1.510{[}0.055{]}(6)+                                  & 0.943{[}0.079{]}(2)+                                  & \cellcolor[HTML]{C0C0C0}\textbf{0.872{[}0.088{]}(1)}  \\
UF8                                     & 1.867{[}0.191{]}(4)+ & NA                                                     & 2.791{[}0.884{]}(6)+                                  & 2.487{[}0.093{]}(5)+  & 0.523{[}0.116{]}(3)+                                  & 0.411{[}0.034{]}(2)+                                  & \cellcolor[HTML]{C0C0C0}\textbf{0.284{[}0.033{]}(1)}  \\
UF9                                     & 1.262{[}0.212{]}(4)+ & NA                                                     & 2.841{[}0.682{]}(6)+                                  & 1.783{[}0.231{]}(5)+  & 0.839{[}0.082{]}(3)+                                  & 0.668{[}0.042{]}(2)+                                  & \cellcolor[HTML]{C0C0C0}\textbf{0.584{[}0.061{]}(1)}  \\
UF10                                    & 6.961{[}0.098{]}(5)+ & NA                                                     & \cellcolor[HTML]{C0C0C0}\textbf{1.214{[}0.132{]}(1)-} & 16.88{[}0.032{]}(6)+  & 4.240{[}0.082{]}(4)+                                  & 3.832{[}0.328{]}(3)+                                  & 3.432{[}0.129{]}(2)                                   \\ \hline
+/=/-                                   & 26/0/1               & 19/0/1                                                 & 24/1/2                                                & 25/2/0                & 24/1/2                                                & 18/8/1                                                & -                                                     \\
\multicolumn{1}{l|}{average rank}       & 5.111                & 4.650                                                  & 4.963                                                 & 5.148                 & 3.778                                                 & 2.259                                                 & 1.370                                                 \\ \hline
\end{tabular}
\end{table*}

\subsection{Experimental Results on Large Scale Benchmark Problems}

To verify the scalability of the proposed BS-MOBO algorithm, we evaluate its performance on larger scale problems. In this experiment, we use the same benchmark problem suits (namely, ZDT, DTLZ, F and UF) as in the previous subsection but now with 50-dimensional decision space and a larger budget of 1000 evaluations for each algorithm.

The Kriging-based surrogate-assisted algorithms with sequential evaluations (ParEGO, HypI and SMS-EGO) are not suitable for this setting mainly due to the extremely long running time. In addition to the model-based optimization algorithms with batch evaluations (K-RVEA and MOEA/D-EGO), we also compare our proposed algorithm with two model-free optimization algorithms as baselines: (1) MOEA/D~\cite{zhang2007moea} is a well-known and popular MOEA for solving MOPs and (2) HIGA-MO~\cite{wang2017hypervolume} is a newly proposed multi-objective optimization algorithm using Hypervolume indicator gradient ascent. The gradient-based HIGA-MO method is now only suitable for solving bi-objective problems, so its results on three objective problems are unavailable.

As mentioned in the previous sections, the time complexity of building a Gaussian process model is cubic to the number of training samples. This high computational cost forbids us to use all evaluated solutions to train the Gaussian process model especially when the evaluation budget is not very small. K-RVEA and MOEA/-EGO both have carefully designed strategies to select a fixed number of evaluated solutions for training the surrogate model. For BS-MOBO-GP, at each iteration, we randomly select a subset of evaluated solutions to train the Gaussian process model.

The parameters settings for this experiment are:

\begin{itemize}
    \item Dimension of decision space $n = 50$;
    \item Maximal number of evaluations $N_{FE} = 20n =  1000$;
    \item Number of solutions in the initial dataset $N_I = 500$;
    \item Number of solutions for training the Gaussian process model $N_{GP}$ = 300;
    \item Number of solutions to be evaluated in batch at each iteration for K-RVEA, MOEA/D-EGO, BS-MOBO and its variants $k = 25$, which means there are only 20 iterations ($(1000 - 500)/25 = 20$) during the optimization process;
    \item Population size for all algorithms is $p = 100$;
    \item Parameters for all compared algorithms are the same with their default settings in the literature;
\end{itemize}

\begin{figure*}[t]
\centering
\includegraphics[width = 0.30\textwidth]{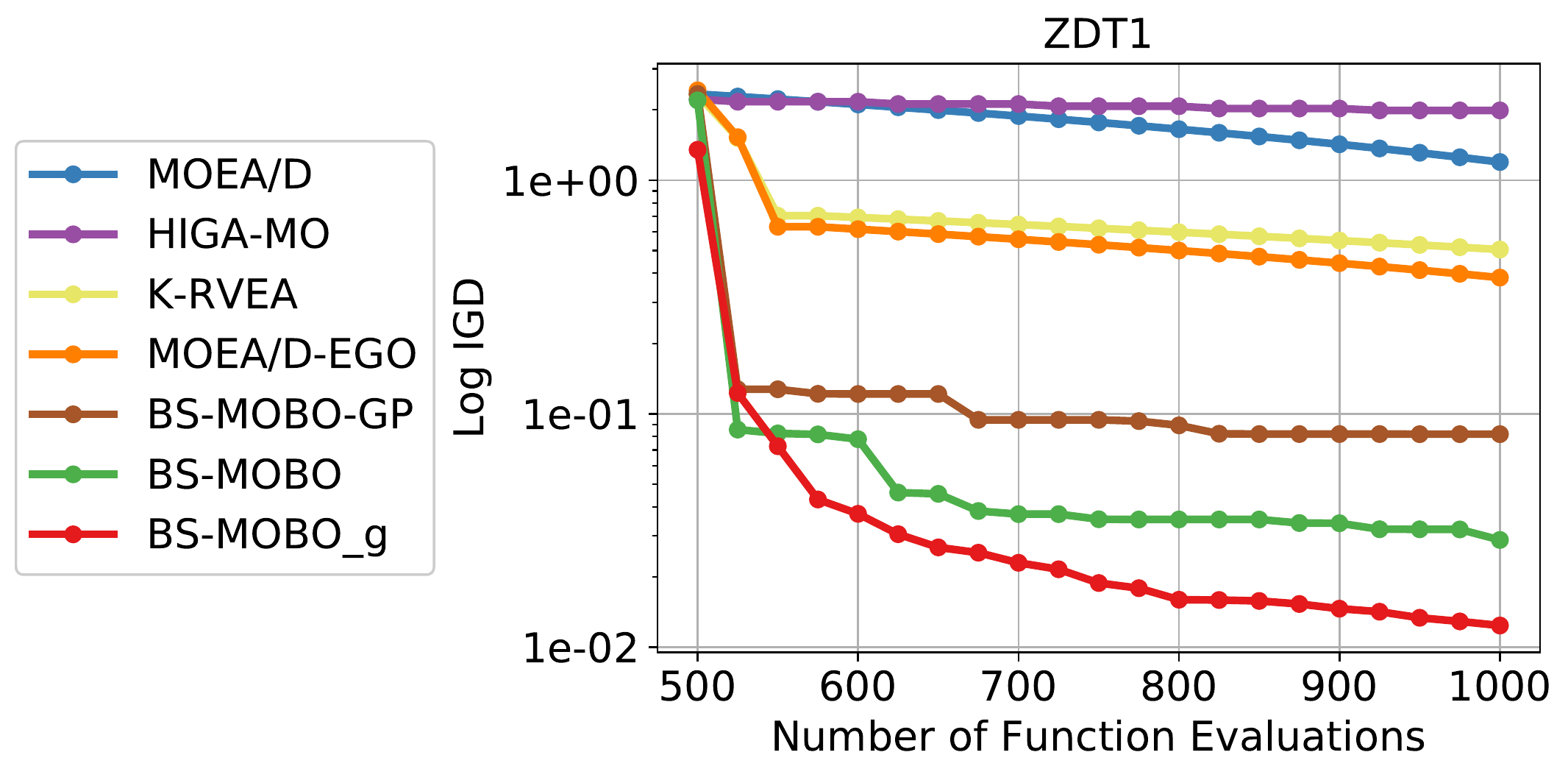}
\includegraphics[width = 0.21\textwidth]{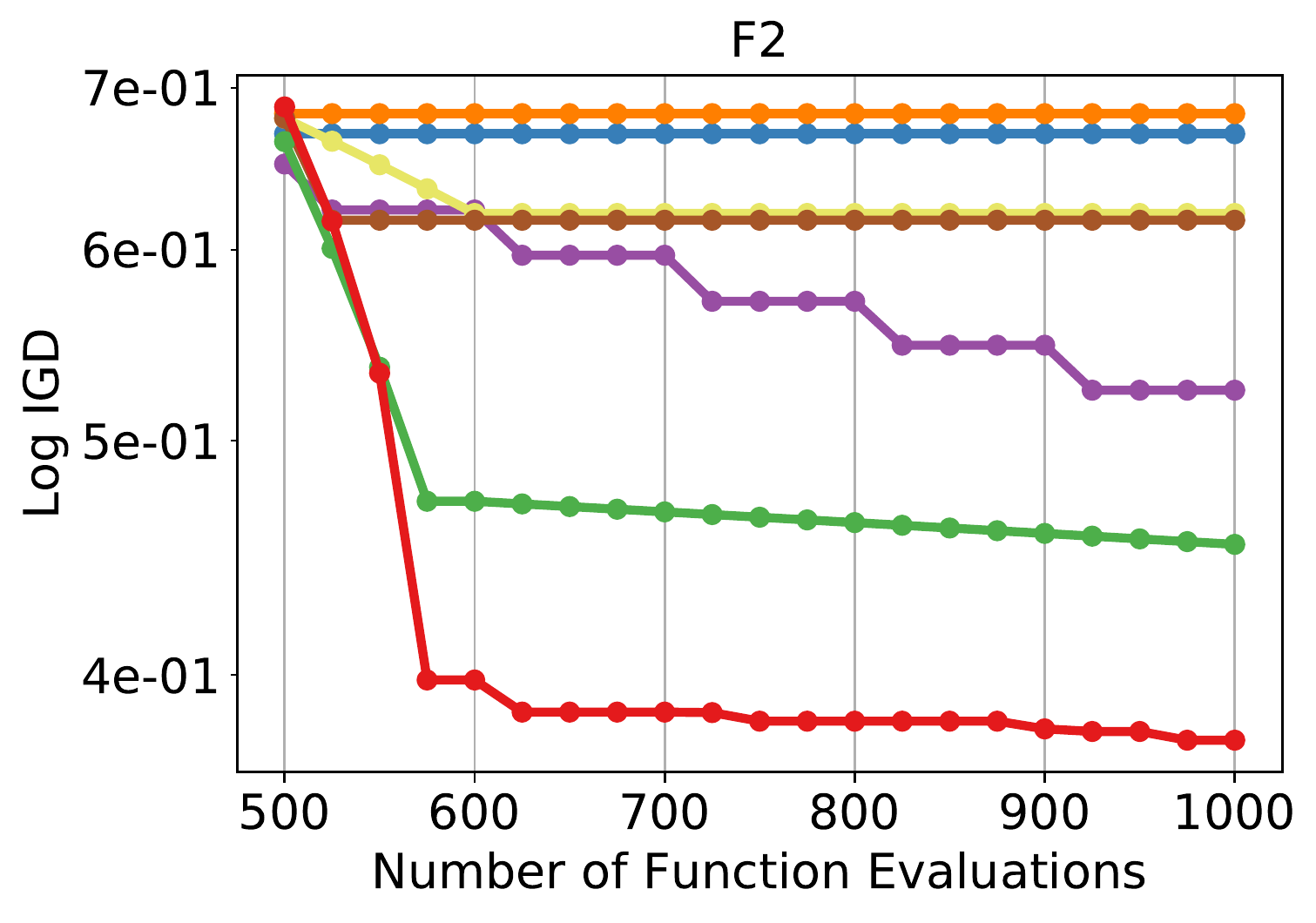}
\includegraphics[width = 0.21\textwidth]{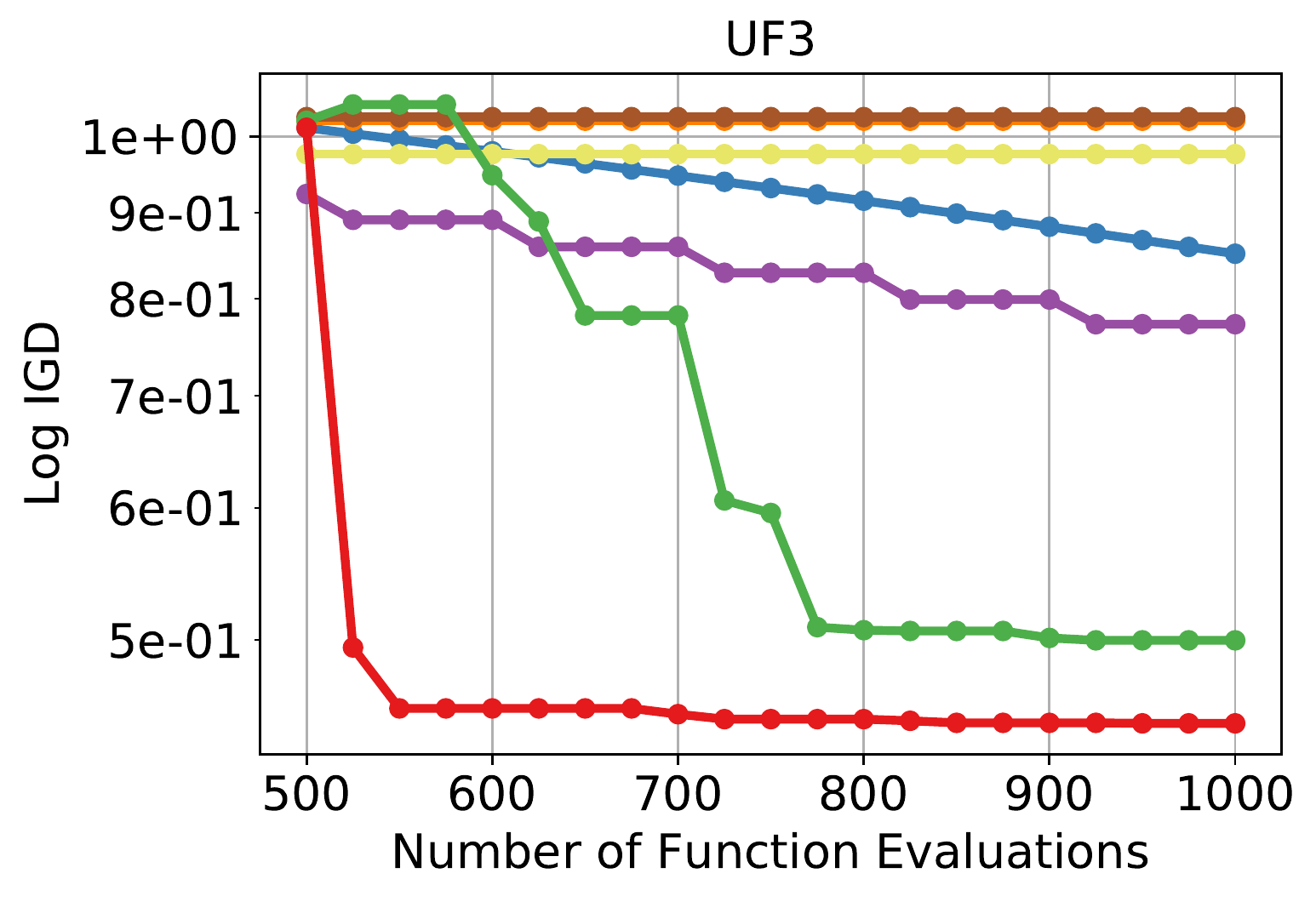}
\includegraphics[width = 0.21\textwidth]{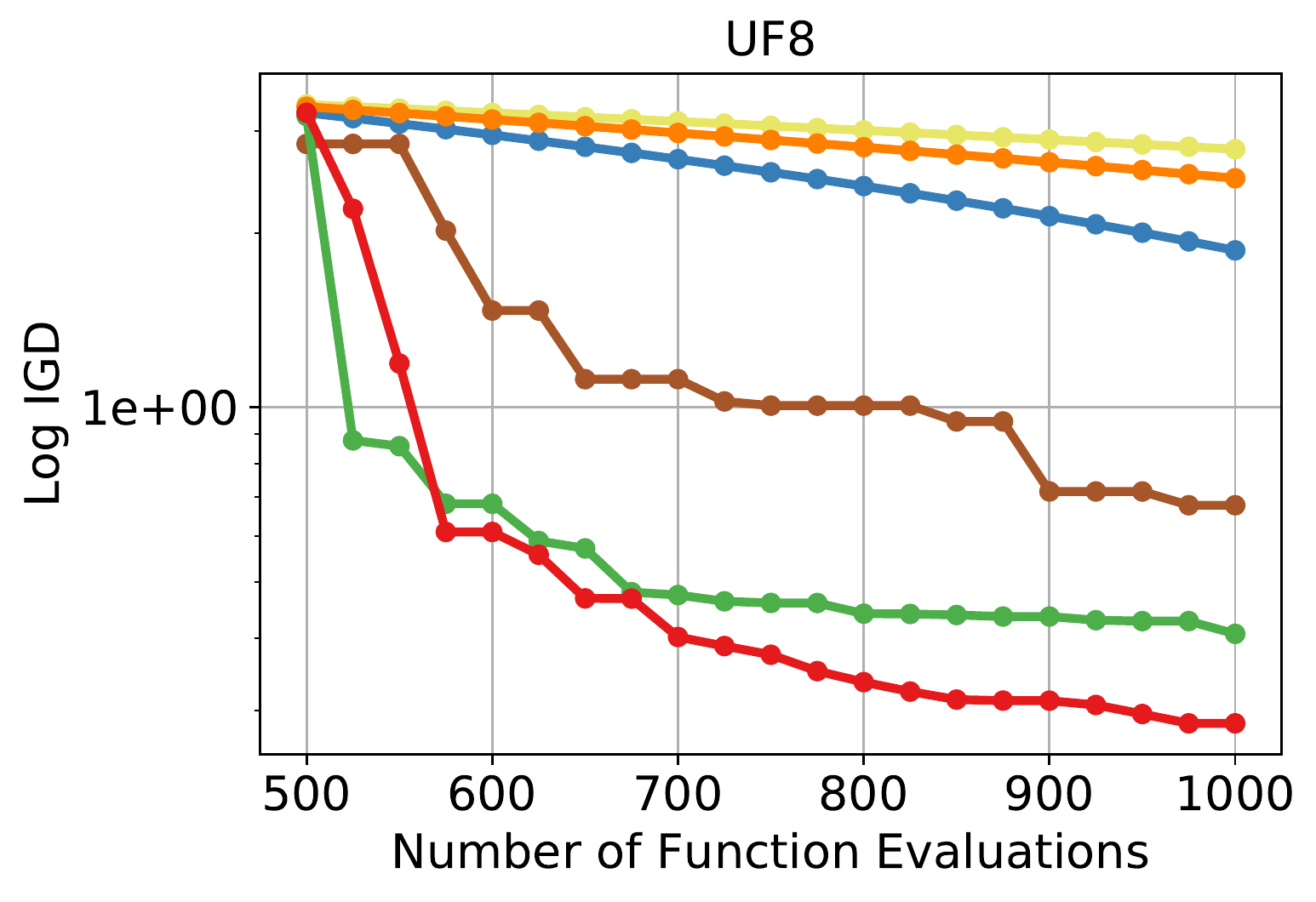}
\caption{Evolutions of the median IGD values obtained by different algorithms versus the number of function evaluations for selected test instances.}
\label{IGD_d50}
\end{figure*}

All algorithms are independently run 25 times, and their performances are compared using the IGD and hypervolume difference metric as mentioned in the previous subsection. The experimental results of the IGD metric are shown in Table.~\ref{results_table_large}. The experimental results of the hypervolume difference metric are presented in the supplementary material.

\subsubsection{The Batch Optimization Framework for Large Scale Problem}
BS-MOBO-GP has a better overall performance compared with K-RVEA, MOEA/D-EGO and those two model-free methods, which confirms the efficiency of our proposed batch optimization algorithm framework for solving large scale problems. It should be noticed that the model-based methods with the Gaussian process model (BS-MOBO-GP, K-RVEA and MOEA/D-EGO) are outperformed by those two model-free methods on some test instances such as ZDT4 and UF1. The Gaussian process model with limited training examples cannot approximate the large scale complicated objective functions well. The model-based methods spend all evaluation budget on exploring the decision space for these problems and therefore have poor convergence performances.

\subsubsection{The Scalability of the Bayesian Neural Network:}
As shown in Table.~\ref{results_table_large}, BS-MOBO (with average rank 2.259) outperforms BS-MOBO-GP (with average rank 3.778) and other algorithms on most test instances. The only difference between BS-MOBO and BS-MOBO-GP is the surrogate models they use. For solving the large scale optimization problems, during the optimization process, a large number of evaluated solutions with high dimensional decision space is available in the dataset. We can easily use all evaluated solutions to train the Bayesian neural network and obtain good predictions for the objective values and uncertainties, which leads to a much better optimization performance.

\subsubsection{The Effect of Gradient Information}
When the gradient information is available, BS-MOBO\_g performs significantly better than BS-MOBO and has the best overall performance with average rank $1.370$. Fig.~\ref{IGD_d50} presents the evolutions of the median IGD values obtained by different algorithms on selected test instances. It is clear that BS-MOBO and BS-MOBO\_g converge faster than other algorithms and BS-MOBO\_g should be the best choice when gradient information is available. HIGA-MO, which is a model-free but gradient-based algorithm, has better performances compared with those Kriging-based surrogate-assisted algorithms on some test instances. However, it is significantly outperformed by BS-MOBO\_g on almost all test test instances. These experimental results confirm that our proposed BS-MOBO algorithm can solve large scale expensive multi-objective optimization problems much better than many existed model-based and model-free algorithms. It is also clear that incorporating available gradient information into the surrogate model building can significantly improve the performance of model-based optimization algorithm.

\subsection{Sensitivity Analysis of the Batch Size}

The batch size $k$ is an important parameter for our proposed BS-MOBO and its variants. When the total evaluation budget is fixed, a larger batch size allows the algorithm to evaluate many candidate solutions in parallel and to have a smaller number of batch evaluations. However, a larger batch size also means a lower update frequency for the surrogate model, which might lead to poor overall performance.

\begin{figure}[h]
\centering
\includegraphics[width = 0.24\textwidth]{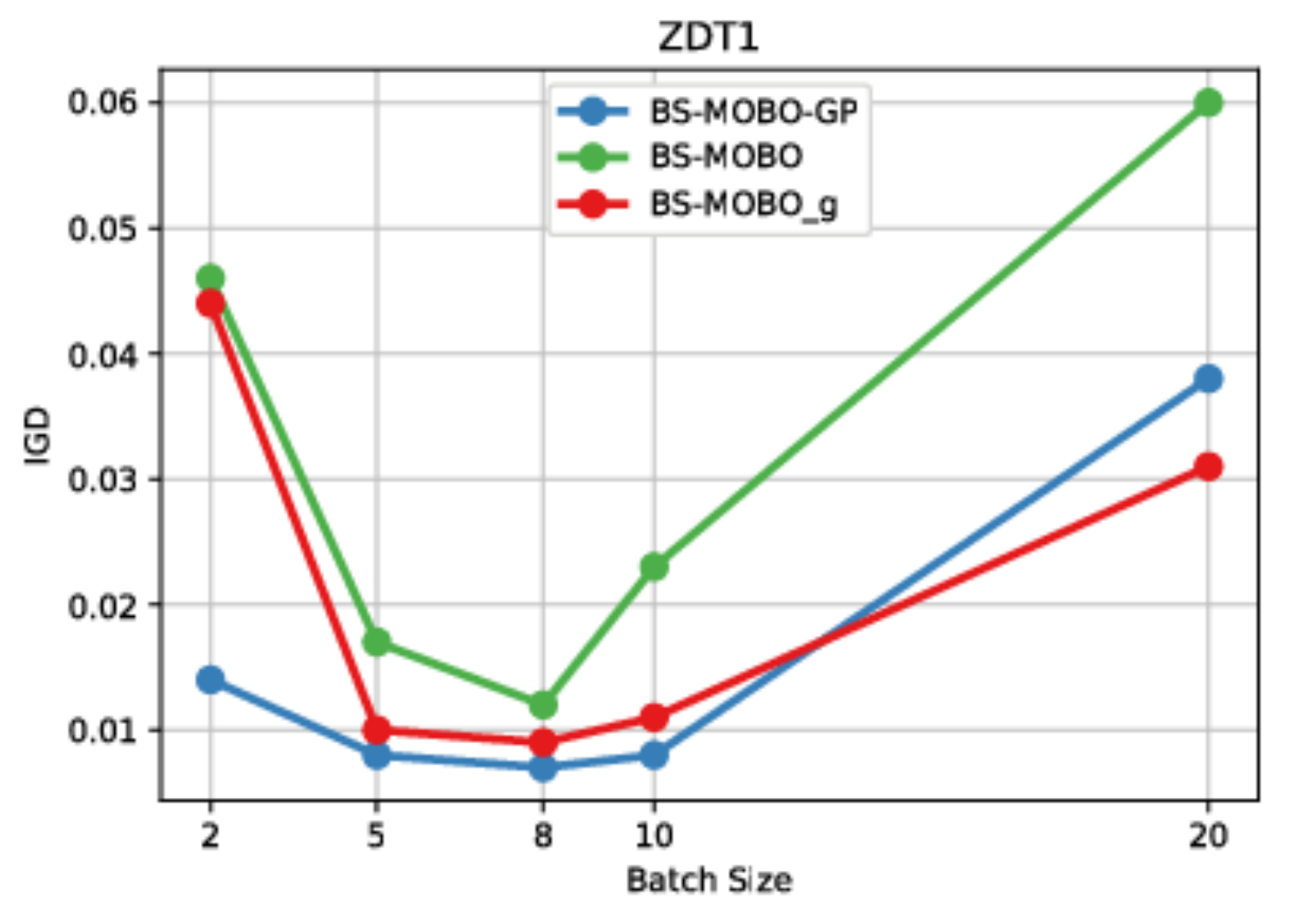}
\includegraphics[width = 0.24\textwidth]{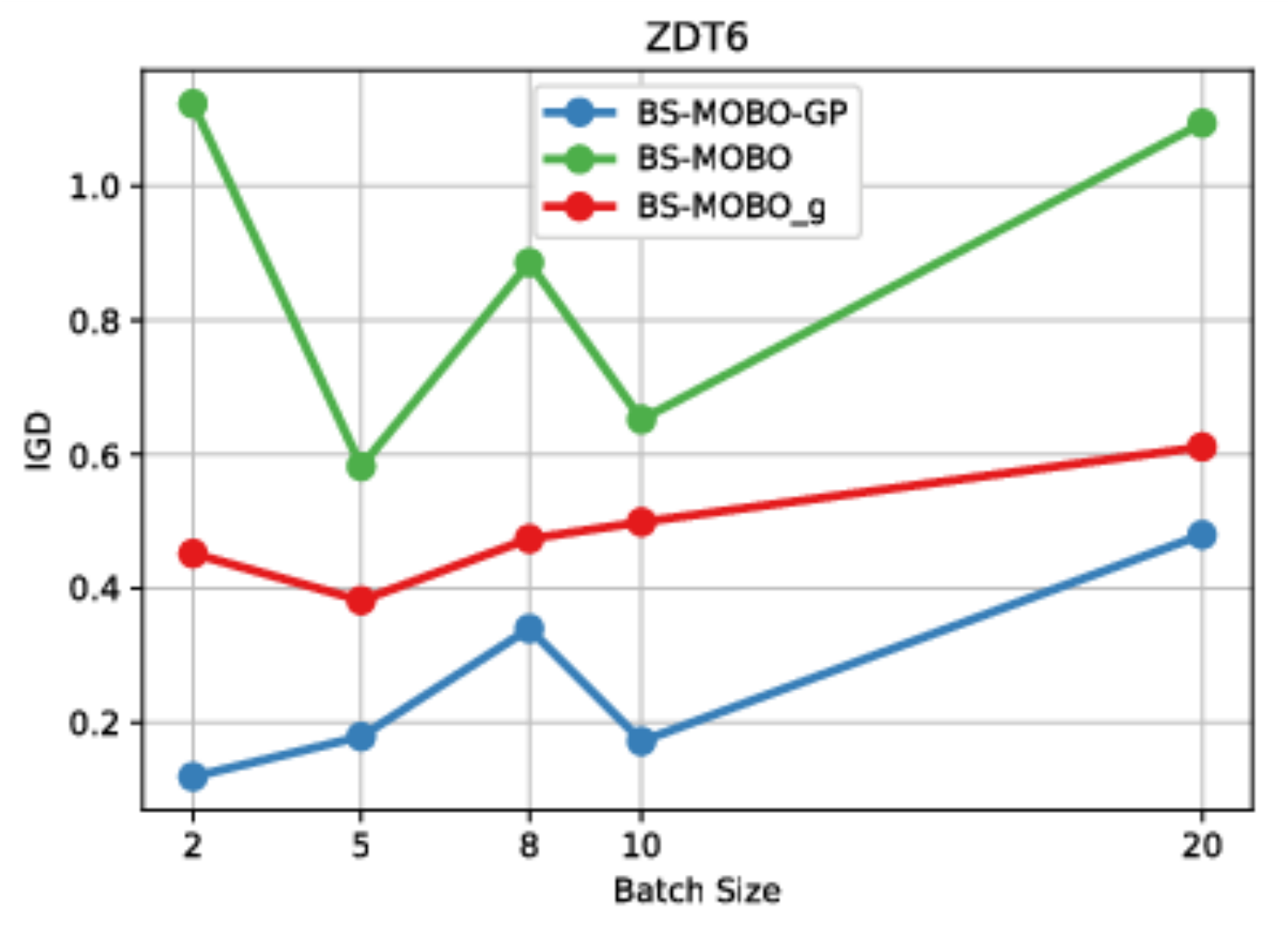}
\\
\includegraphics[width = 0.24\textwidth]{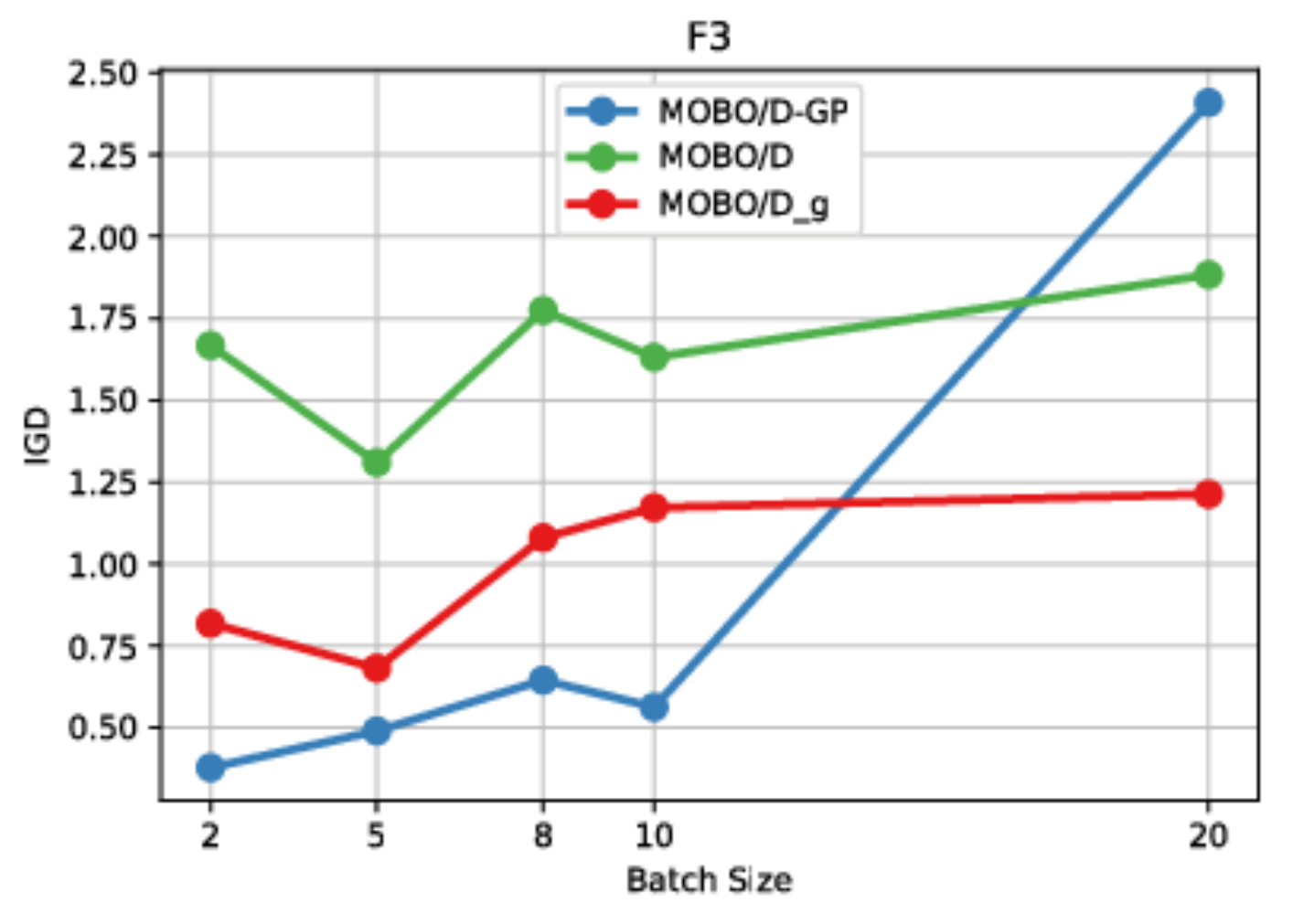}
\includegraphics[width = 0.24\textwidth]{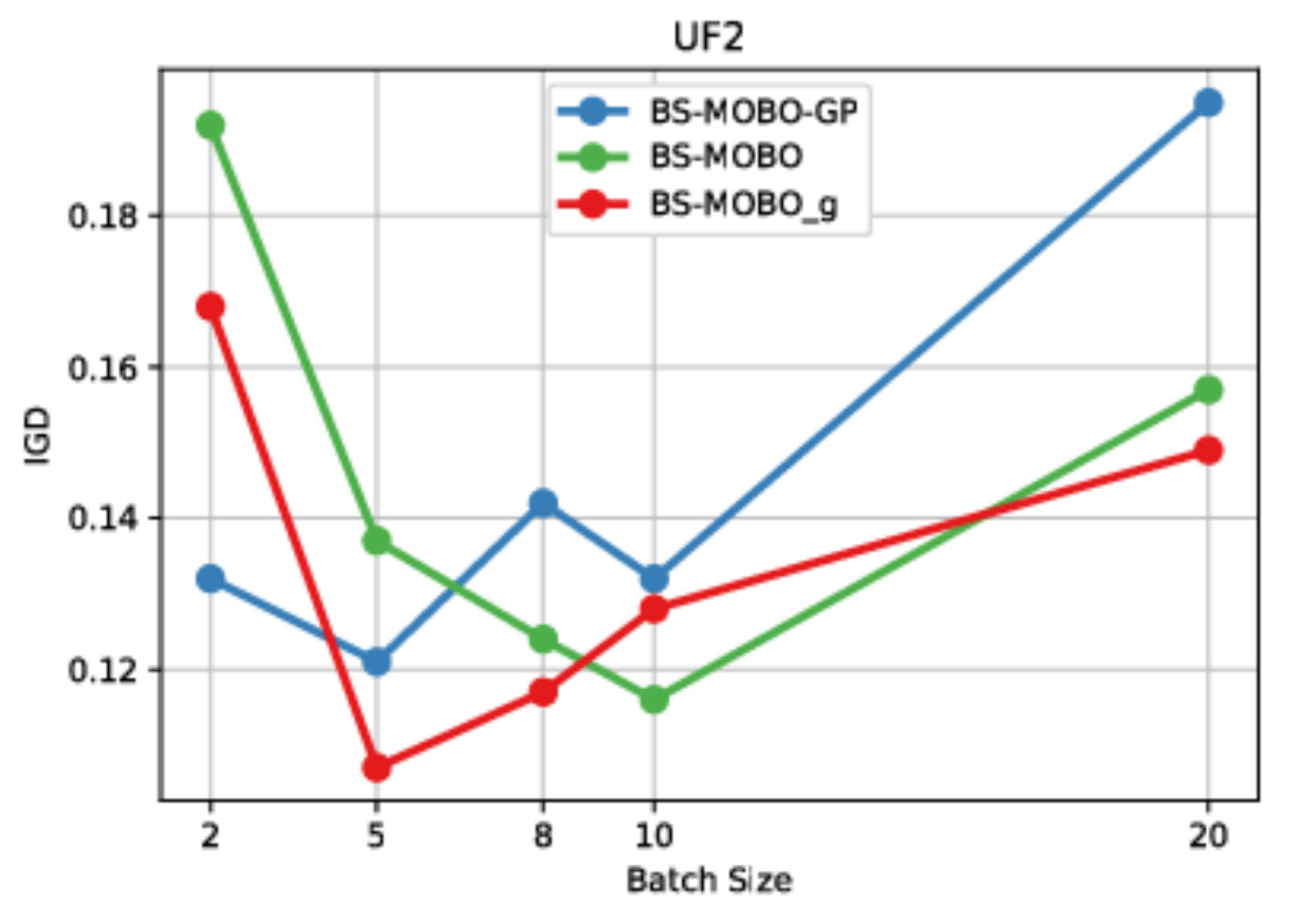}
\caption{The median IGD values obtained by BS-MOBO and its variants on selected 8 dimensional test instances with different batch sizes.}
\label{Batch_size}
\end{figure}

Fig.~\ref{Batch_size} shows a sensitivity analysis of the batch size for BS-MOBO and its variants on four selected 8-dimensional test instances. There is no single best choice of batch size for different algorithms on different problems. It is worth mentioning that, in single objective Bayesian optimization, batch evaluations usually have inferior performance compared with sequential evaluation~\cite{desautels2014parallelizing}. Larger batch size can lead to a poorer performance when the goal is to find a single best solution for single objective function. But it is not the case for model-based multi-objective optimization algorithms.

In BS-MOBO, we need to find a set of solutions for parallel evaluation to balance the convergence and diversity at each iteration. If the batch size is too small (e.g., 2), we will not have enough solutions to fully explore the estimated Pareto front given by the surrogate models, and some promising solutions cannot be evaluated. On the other hand, if the batch size is too large (e.g., 20) when the total evaluation budget is small and fixed (e.g., 100), the number of batch evaluations would be very small (e.g., 100/20 = 5). In other words, the surrogate models will be only updated a few times during the optimization process, which might lead to poor approximation and inferior overall performance. To achieve the best performance, the batch size should be changed adaptively for different problems and in different optimization stage. However, there is no explicit rule to set the adaptive schedule of the batch size for BS-MOBO and other batch-based algorithms such as MOEA/D-EGO~\cite{zhang2010expensive} and K-RVEA~\cite{chugh2016surrogate}. Dynamically adapting the batch size is an important future work to further improve the performance of bathed multi-objective optimization algorithms.

\subsection{Space Engineering: Spacecraft Trajectory Optimization}

In this subsection, we compare the performance of our proposed BS-MOBO and its variants with different algorithms on a real-world spacecraft trajectory optimization problem. The trajectory optimization problem is challenging and crucial to the whole mission design~\cite{izzo2010global}.

The problem we deal with is the Rosetta space mission problem. This mission aims at sending a spacecraft to the comet 67P/Churyumov-Gerasimenko. We treat it as a bi-objective optimization problem as in~\cite{fliege2016method}. The first objective is the total $\Delta V$ which measures the amount of impulse and also indicates the amount of fuel used in the mission. The second objective is the squared total travel time.  This problem has total 22 decision variables and their gradient information can be easily obtained. The details of this Rosetta problem can be found in the paper~\cite{fliege2016method} as well as on the European Space Agency (ESA) website\footnote{https://www.esa.int/gsp/ACT/projects/gtop/rosetta.html}.

\begin{figure}[H]
    \centering
    \includegraphics[width= 0.50\textwidth]{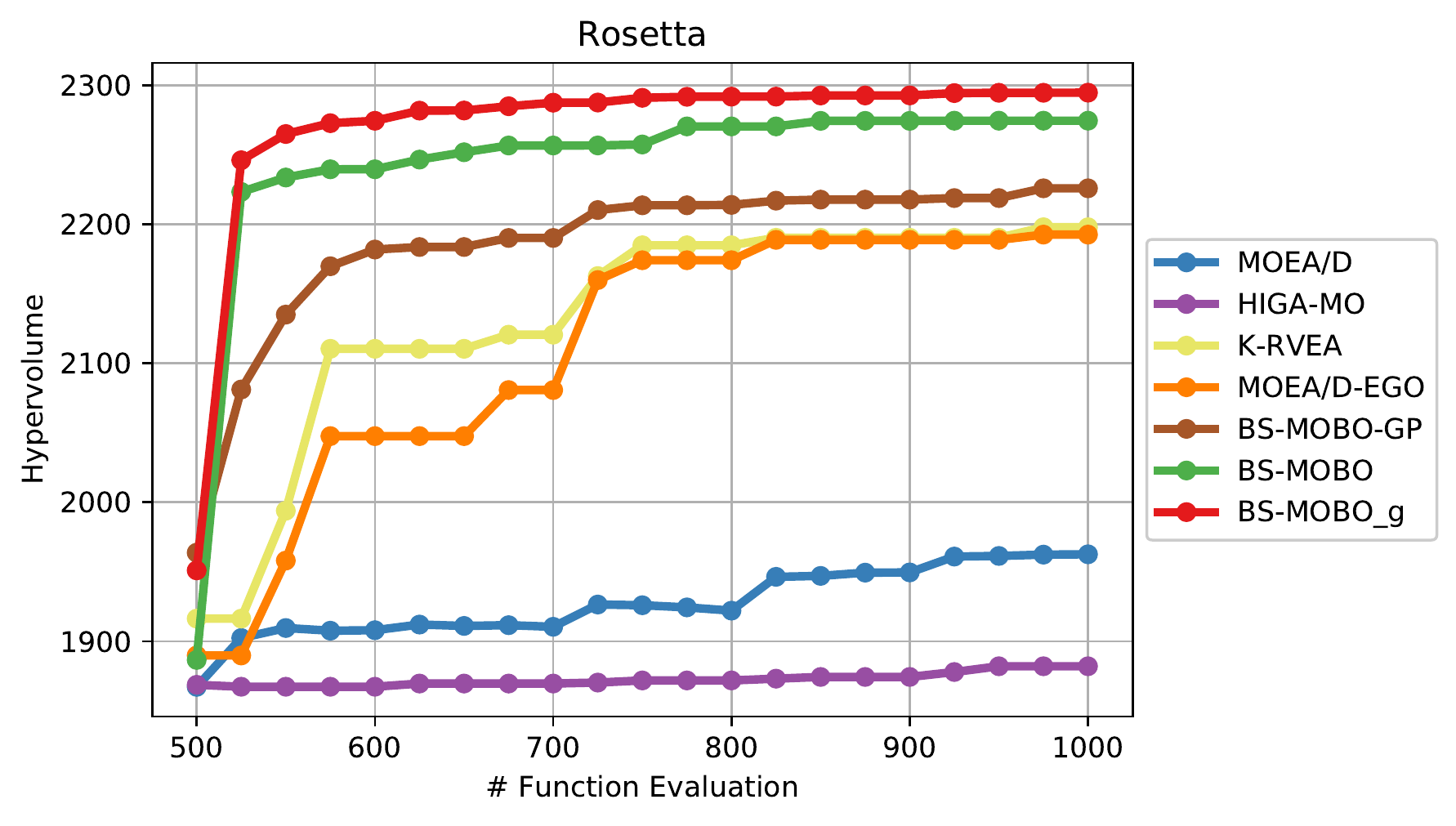}
    \caption{\label{fig:rosetta} Evolution of the median Hypervolume values obtained by different algorithms versus the number of function evaluations on the rosetta problem.}
\end{figure}

\begin{figure}[H]
    \centering
    \includegraphics[width= 0.50\textwidth]{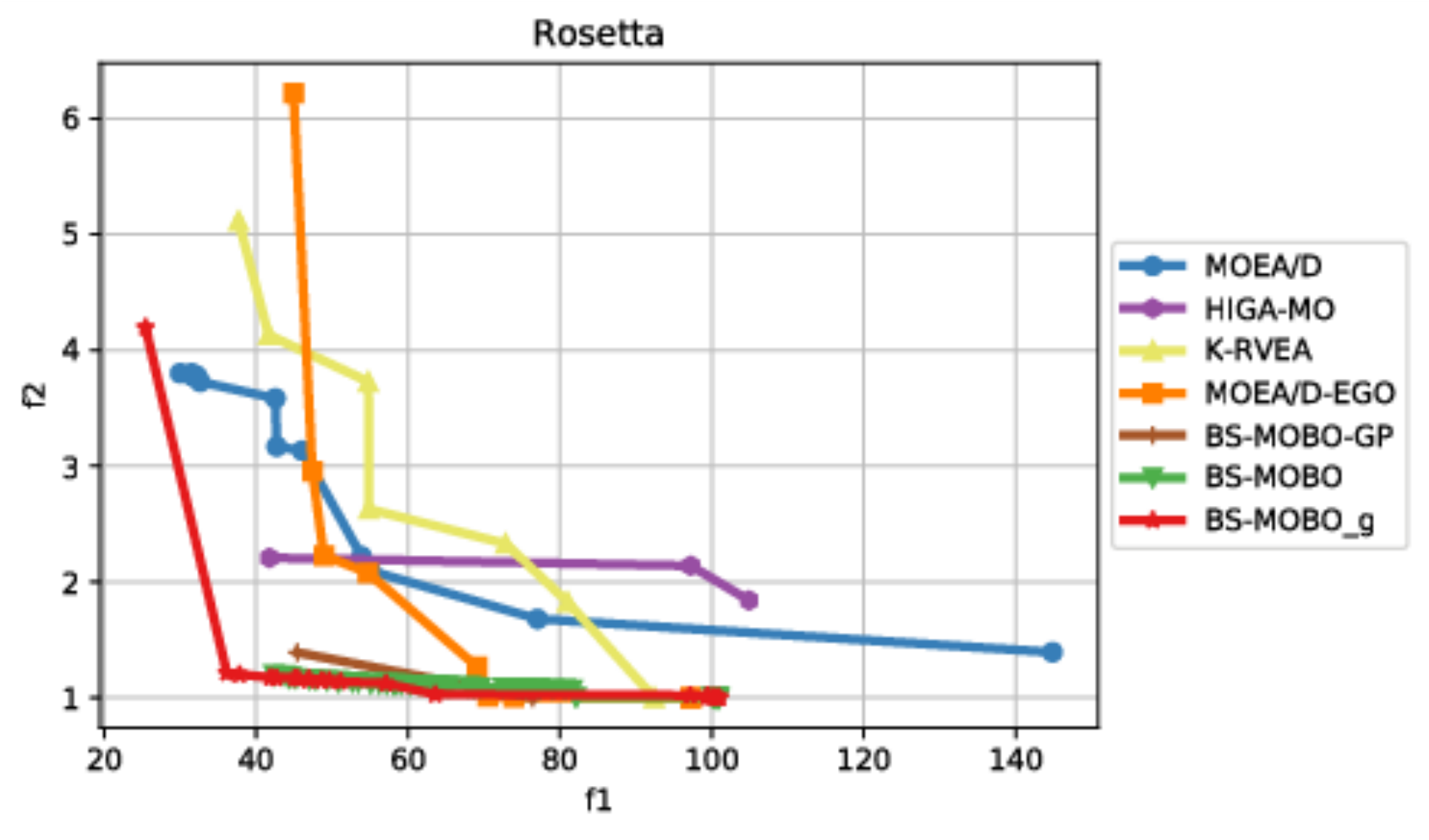}
    \caption{\label{fig:pf_rosetta} The approximated Pareto front obtained by different algorithms on the Rosetta problem.}
\end{figure}

We independently run each algorithm 10 times. Fig.~\ref{fig:rosetta} shows the optimization performances of different algorithms. Since we do not know the ground truth Pareto front of this problem, we report the hypervolume with reference point $[500,8]$ as the performance metric. It is clear that BS-MOBO and BS-MOBO\_g have significantly better performance and BS-MOBO\_g should be the best choice when the gradient information is available. Fig.~\ref{fig:pf_rosetta} presents the best approximated Pareto front obtained by different algorithms. The solutions obtained by BS-MOBO and BS-MOBO\_g dominate most solutions obtained by other algorithms, which again confirm their superior performances.

\section{Conclusion}
In this paper, we have proposed BS-MOBO, a batched scalable multi-objective Bayesian optimization algorithm, for solving large scale expensive multi-objective optimization problems. Our proposed algorithm builds scalable Bayesian neural network models to estimate the values and uncertainties of each optimization objective. Powered with the Sobolov training method, the models can incorporate available gradient information and provide good prediction performance for solutions in the high dimensional decision space with a limited number of evaluated solutions. In addition, we have also proposed an efficient batched multi-objective acquisition function to generate promising solutions for parallel evaluations, which is important for many real-world applications.

We have compared the performance of the proposed BS-MOBO and its variants with many state-of-the-art surrogate-assisted MOEAs as well as two typical model-free optimization algorithms. It is clear that BS-MOBO can achieve significantly better performance on various large scale benchmark problems and a real-world problem with high dimensional decision space. The experimental results also confirm that incorporating available gradient information is crucial for obtaining a promising optimization performance for large scale expensive optimization problem.

In the proposed BS-MOBO, the batch size is fixed (5 for 8-dimensional problems, and 25 for 50-dimensional problems) during the whole optimization process for all problems. However, as shown in the experimental study, the optimal setting for the batch size is problem-dependent. In addition, different batch sizes might also be beneficial for different optimization stages in the whole optimization process. How to adaptively change the batch size is an important and interesting research topic in our future work.

\appendices





\ifCLASSOPTIONcaptionsoff
  \newpage
\fi



\bibliographystyle{ieeetr}
\bibliography{citation}

\end{document}